\documentclass[journal]{IEEEtran}

\ifCLASSINFOpdf
\else
\fi

\usepackage{setspace}

\usepackage{times}
\usepackage{subfig}
\usepackage{graphicx}
\usepackage{amsmath,amsthm}
\usepackage{amssymb,wasysym}
\usepackage{mathrsfs}
\usepackage{cite}
\usepackage{xcolor}
\usepackage{url}
\usepackage[lined,linesnumbered,ruled]{algorithm2e}
\usepackage{wrapfig}
\usepackage{verbatim}
\usepackage{dsfont}
\usepackage{algorithmic}
\usepackage{multirow}

\ifodd 0
\newcommand{\rev}[1]{{\color{green}#1}} 
\newcommand{\newrev}[1]{{\color{blue}#1}} 
\newcommand{\needrev}[1]{{\color{red}#1}} 
\newcommand{\nextrev}[1]{{\color{orange}#1}} 
\newcommand{}[1]{{\color{pink}#1}} 
\else
\newcommand{\rev}[1]{#1}
\newcommand{\newrev}[1]{#1} 
\newcommand{\needrev}[1]{#1} 
\newcommand{\nextrev}[1]{#1} 
\fi

\begin{document}

\newcommand{\name}{IC3M\xspace}

\title{\name: In-Car Multimodal Multi-object Monitoring for Abnormal Status of Both Driver and Passengers}

\author{Zihan Fang, Zheng Lin, Senkang Hu, Hangcheng Cao, Yiqin Deng,~\IEEEmembership{Member,~IEEE}, \\
Xianhao Chen,~\IEEEmembership{Member,~IEEE} and Yuguang Fang,~\IEEEmembership{Fellow,~IEEE}

\thanks{Z. Fang, S. Hu, H. Cao, Y. Deng and Y. Fang are with the Department of Computer Science, City University of Hong Kong, Kowloon, Hong Kong SAR, China (e-mail: zihanfang3-c@my.cityu.edu.hk; senkang.forest@my.cityu.edu.hk; hangccao@cityu.edu.hk; yiqideng@cityu.edu.hk; my.fang@cityu.edu.hk).}
\thanks{Z. Lin and X. Chen are with the Department of Electrical and Electronic Engineering, The University of Hong Kong, Pok Fu Lam, Hong Kong, China (e-mail: linzheng@eee.hku.hk; xchen@eee.hku.hk).}
}

\markboth{Journal of \LaTeX\ Class Files,~Vol.~14, No.~8, August~2015}%
{Shell \MakeLowercase{\textit{et al.}}: Bare Advanced Demo of IEEEtran.cls for IEEE Computer Society Journals}

\maketitle

\begin{abstract}
Recently, in-car monitoring has emerged as a promising technology for detecting early-stage abnormal status of the driver and providing timely alerts to prevent traffic accidents. Although training models with multimodal data  enhances the reliability of abnormal status detection, the scarcity of labeled data and the imbalance of class distribution impede the extraction of critical abnormal state features, significantly deteriorating training performance. 
Furthermore, missing modalities due to environment and hardware limitations further exacerbate the challenge of abnormal status identification.
More importantly, monitoring abnormal health conditions of passengers, particularly in elderly care, is of paramount importance but remains underexplored.
To address these challenges, we introduce our \name, an efficient camera-rotation-based multimodal framework for monitoring both driver and passengers in a car. Our \name comprises two key modules: an adaptive threshold pseudo-labeling strategy and a missing modality reconstruction. The former customizes pseudo-labeling thresholds for different classes based on the class distribution, generating class-balanced pseudo labels to guide model training effectively, while the latter leverages cross-modality relationships learned from limited labels to accurately recover missing modalities by distribution transferring from available modalities. 
Extensive experimental results demonstrate that \name outperforms state-of-the-art benchmarks in accuracy, precision, and recall while exhibiting superior robustness under limited labeled data and severe missing modality.
\end{abstract}


\IEEEpeerreviewmaketitle

\vspace{-0.2cm}
\section{Introduction} \label{sec:introduction}
Traffic accidents result in a significant number of fatalities and injuries each year.
\newrev{Association for Safe International Road
Travel shows that nearly 1.35 million people die in road crashes every year, or an average of 3,287 deaths per day~\cite{Association}. At this rate, traffic accidents is ranked as the fifth leading cause of death by 2030~\cite{WHO}.}
According to~\cite{dingus2016driver}, one of the leading cause to traffic accidents is the abnormal status of driver (e.g., distraction~\cite{jegham2019mdad}, stress~\cite{rastgoo2018critical}, fatigue~\cite{zhang2016traffic}, and health emergency~\cite{hamza2020monitoring}), accounting for about 87.7\% of the overall road crashes.
Those abnormal statuses greatly impair driver's hazard perception and decision-making ability~\cite{nvemcova2020multimodal}, hence, the early detection of a driver's abnormal status is imperative to avoid traffic accidents.
Therefore, the effective in-car monitoring systems that can identify abnormal status at early stage and reminder a driver in advance have acquired immense popularity recently~\cite{kaplan2015driver, juncen2023mmdrive, pakdamanian2021deeptake}.

\begin{figure}[t] 
\centering
\includegraphics[width=\linewidth]{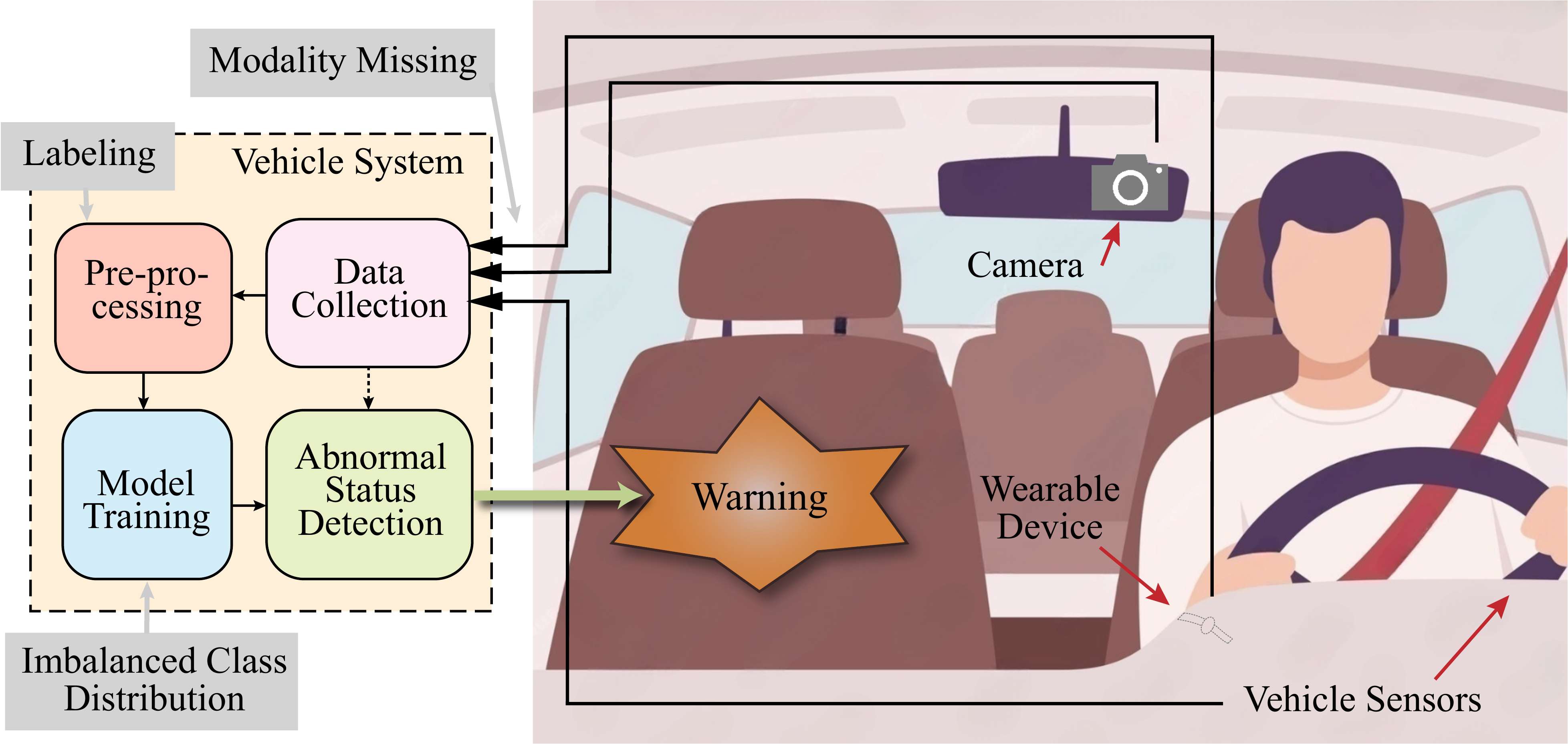}
\caption{The typical scenario of driver monitoring with multiple sensing modalities.}
\label{fig:driver_challenge}
\vspace{-2ex}
\end{figure}

To achieve a comprehensive and reliable monitoring system,
different sensing modalities need to be leveraged collaboratively as a single modality alone is not always informative for status detection~\cite{du2020vision}.
Cameras could provide precise records for facial expression (e.g., eye blinking and yawning) but perform poorly in insufficient lighting and occlusion~\cite{alioua2014driver, sigari2013driver, yang2018car}, whereas wearable sensors provide robust monitoring for physiological features (e.g., respiration rates and heartbeats) against adverse conditions~\cite{ghorbani2023self}.
By integrating diverse data from various sensing modalities, the system captures complementary information from different perspectives, significantly enhancing the reliable perception in a vehicle.
In general, as depicted in Fig.~\ref{fig:driver_challenge}, a vehicle trains a model to recognize a driver's abnormal status with multimodal data and then notifies the driver to take necessary actions in advance.

Nevertheless, performing in-car status monitoring upon multimodal data encounters several major challenges \needrev{on informative representation learning.}
First, the large amount of well-labelled data is usually difficult to acquire.  
Since the evaluation of abnormal status always heavily relies on a user's feeling with vague definition, the unreliable and subjective evaluation leads to label inconsistency and omission~\cite{nvemcova2020multimodal, taamneh2017multimodal}.
However, current in-car monitoring frameworks~\cite{du2020vision, pakdamanian2021deeptake, yang2023aide} are typically based on supervised learning which depends on vast datasets with high-quality labelled data to learn the robust feature representations~\cite{hu2024collaborative,lin2024efficient,peng2024sums,yuan2024satsense,hu2023towards,zhang2024satfed,lin2022channel,hu2023adaptive}.
\nextrev{The limited labeled data impedes model ability to discover the underlying relationships of feature representations, thus posing a significant barriers to extract informative feature representations efficiently.}
Meanwhile, recognizing abnormal status with the imbalanced data quantity in different classes is challenging.
{The unpredictability to foresee and record abnormal events leads to a smaller amount of data for abnormal status compared to normal status~\cite{bao2024bmad}, while the artificially generated data may not fully capture the wide range of possible patterns~\cite{ghorbani2024personalized}.}
The imbalanced class distribution for model training makes it hard to learn characteristics of abnormal status, causing model to misclassify data as normal status, which may jeopardize the lives of people in the car without prompt detection of abnormal status.

Moreover, the acquisition of full-modality data in a vehicle system is also a challenge as sensing limitations, vibration of a vehicle, lighting condition and occlusion all contribute to the {possible data lost}~\cite{ma2021smil}. 
Unfortunately, the existing multimodal learning frameworks assume that the complete data samples with all modalities are always available.
\nextrev{The lack of auxiliary data handicaps the training model to distinguish similar activities like blinking eyes or fainting and feeling hot or under stress, hence the reliability of informative representation learning could not be guaranteed.}
Furthermore, the substantial differences in dimensions and patterns among modalities also cause distinct representations in feature space~\cite{wang2020multimodal}, posing significant challenge to recover the important details for the missing modality.  
Although continuous monitoring of the driver is paramount, prompt responses to abnormal health conditions (e.g., \needrev{cardiovascular disease and stroke}) of other passengers is also non-trivial.
\nextrev{However, the health monitoring of passengers has been relatively overlooked in current in-car monitoring system compared to driving safety.}
Due to the long-time driving, driving under time pressure, and nighttime driving, the abnormal status of passengers may not be noticed by others in time~\cite{nvemcova2020multimodal}, putting them, particularly the elderly, at a greater risk to their health as cardiovascular disease accounts for more than 40\% of total deaths and \rev{85\% of deaths due to chronic diseases occurs in those older than 70 years~\cite{wang2016global, fuster2018somatic}.}
As a result, continuously monitoring the status of both driver and other passengers in a vehicle is highly desired for driving safety and elderly care as well.

To simultaneously monitor the abnormal status of a driver and passengers, rotating camera emerges as a viable solution {by continuously changing its viewing angles to provide broader coverage~\cite{cho2011sector, lane1998tracking, perera2006multi}.} Therefore, mounting a rotating camera on the rear-view mirror enables efficient in-car detection of driver and passengers, guaranteeing continuous detection of them within its expanded field of view. 
Despite the promising potential of rotating cameras, the implementation of rotating cameras significantly degrades the reliability and efficiency of multi-object monitoring:
{\romannumeral1)} \nextrev{Turning the angle of the camera introduces noise, resulting in low-quality data collected.} 
{\romannumeral2)} Camera rotation causes severe modality missing when passengers are out of covering range.
{\romannumeral3)} \nextrev{The distorted facial images from different angles and perspectives enlarge the distribution gap between visual and physiological feature representations, necessitating effective alignment of different modalities for modality reconstruction.}

In-car multimodal monitoring has been extensively investigated for several years. However, none of the existing multimodal sensing systems~\cite{du2020vision, pakdamanian2021deeptake, yang2023aide} is capable of \nextrev{learning informative representations with limited high-quality labels and severe missing modalities to monitor multiple objects concurrently.}
Moreover, existing efforts only focus on monitoring the driver whereas the status of other passengers are neglected, failing to fully utilize the perception capabilities of a vehicle to enhance the overall safety.
\nextrev{Hence, it is imperative to design a multi-object monitoring system that effectively learns informative representations with limited labels under severe modality missing.}

In this paper, we propose a cutting-edge in-car monitoring system, named \underline{i}n-\underline{c}ar \underline{m}ultimodal \underline{m}ulti-object \underline{m}onitoring (\name), to recognize the abnormal status at the early stage {of critical health conditions} for both driver and passengers. 
\name comprises two well-designed modules, namely, adaptive threshold pseudo-labeling and missing modality reconstruction. 
The adaptive threshold training is designed to generate high-confidence data (i.e., unlabeled data with confidence higher than the threshold) through adaptive threshold adjustment to achieve more balanced class distribution for model training. 
As for the low confidence data, we introduce it into model training through \needrev{contrastive learning}, facilitating feature extraction for informative feature representations and thereby accelerating convergence.
To keep the model performance when some sensing data is missing, we propose a novel modality reconstruction network to restore modality-specific features across the distribution mapping from the available modality to the missing modality. 
\rev{The tailored \needrev{meta learning framework} is utilized for generalization enhancement with limited labels.}
We summarize our main contributions as follows.
\begin{itemize}
    \item We propose an adaptive threshold pseudo-labeling scheme to dynamically adjust confidence threshold based on class quantity with the incorporation of low-confidence unlabeled data in model training.
    \item We design a feature reconstruction network to learn the distribution mapping through limited labels according to cross-modality relationships.
    \item \rev{Our proposed framework is a more general model which is not limited in this in-car monitoring scenario, and could be applicable to other multimodal systems.}
    \item We conduct extensive experiments to illustrate that \name achieves higher \needrev{detection accuracy} and convergence speed than state-of-the-art benchmarks, demonstrating the effectiveness of our well-designed adaptive threshold pseudo-labeling and modality reconstruction modules.
\end{itemize}

This paper is organized as follows. Sec.~\ref{sec:motivation} motivates the design of \name\ by revealing the limitations in current in-car status monitoring system. Sec.~\ref{sec:system_design} presents the system design of \name. Sec.~\ref{sec:implementation} introduces system implementation, and experimental setup, followed by performance evaluation in Sec.~\ref{sec:evaluation}. Related works and technical limitations are discussed in Sec.~\ref{sec:related_work}. Finally, conclusions and future remarks are presented in Sec.~\ref{sec:conclusion}.

\vspace{-0.2cm}
\section{Background and Motivation} \label{sec:motivation}
To better motivate the design of \name, we first provide extensive pilot measurements to elaborate on the challenges that limited labelled data and modality missing pose on in-car status monitoring. \nextrev{Next, we analyze the extra difficulties associated with multi-object monitoring in a vehicle and exhibit the superiority of rotating camera as a potential solution.}

\vspace{-0.2cm}
\subsection{Limited Labels with Imbalanced Class for Model Training} \label{sec:mtv_label}
\begin{figure}[t]
\centering
\subfloat[Accuracy \label{fig:label_quantity_accuracy}]{
    \includegraphics[width=0.50\linewidth]{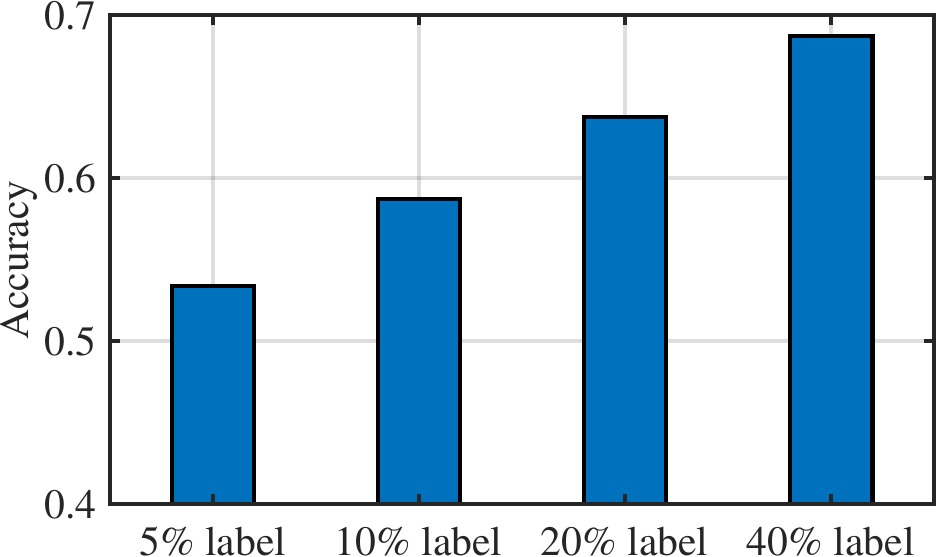}
}
\subfloat[Convergence \label{fig:label_quantity_convergence}]{
    \includegraphics[width=0.50\linewidth]{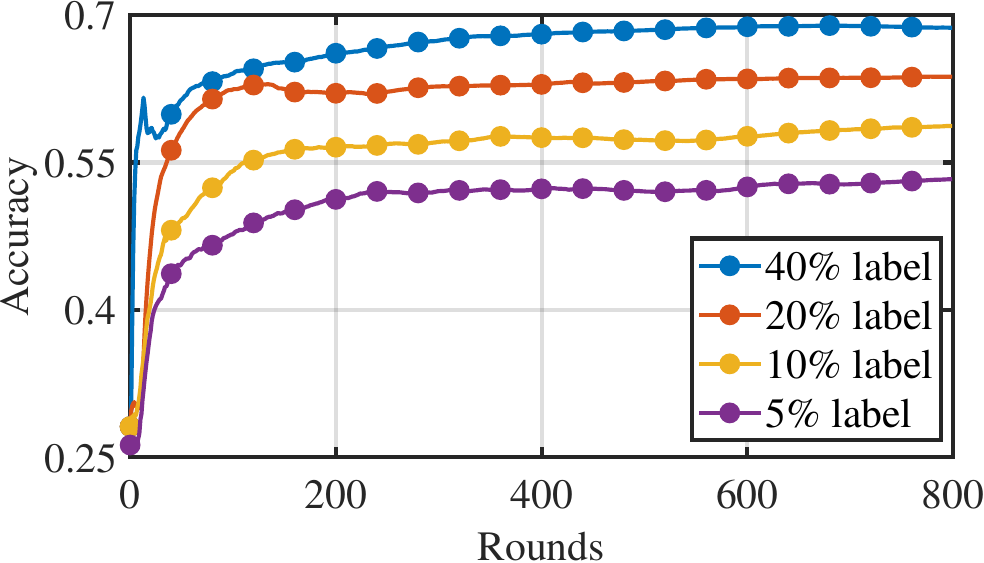}
}
    \caption{Performance comparison with different proportions of labelled samples in the total training samples.}
    \label{fig:label_quantity}
    \vspace{-2ex}
\end{figure}

\begin{figure}[t]
\centering
\subfloat[Performance \label{fig:label_class_accuracy}]{
    \includegraphics[width=0.50\linewidth]{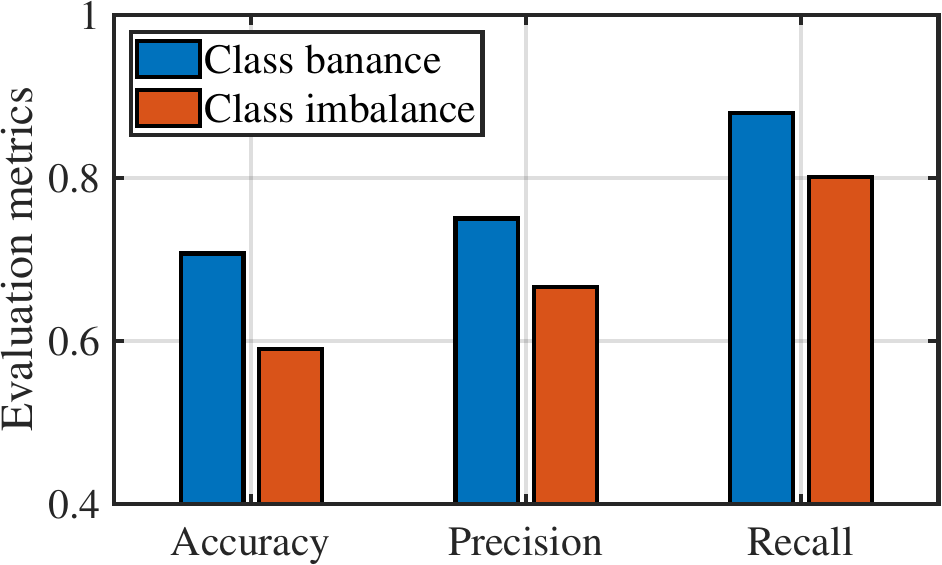}
}
\subfloat[Convergence \label{fig:label_class_convergence}]{
    \includegraphics[width=0.50\linewidth]{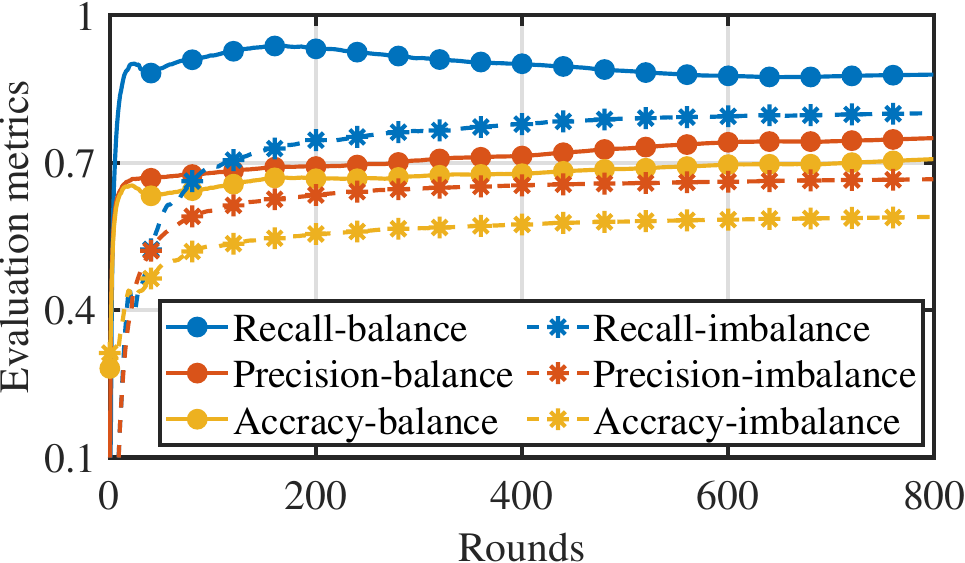}
}
    \caption{Performance comparison with balanced and imbalanced training dataset with 90\% data from normal class and 10\% data from abnormal status.}
    \label{fig:label_class}
    \vspace{-2ex}
\end{figure}

State-of-the-art in-car monitoring frameworks~\cite{du2020vision, pakdamanian2021deeptake, yang2023aide} are typically based on supervised learning with the availability of a large amount of labelled data.
However, in real-world scenarios, only a small amount of well-labelled data may be available as 
the unreliable subjective evaluation of current status results in discrepancies and omissions between labels and actual situations~\cite{nvemcova2020multimodal, taamneh2017multimodal}, severely degrading the performance of the model training.
{Moreover, the limited availability of abnormal status brings the unbalanced class distribution for model training.
This means that the model will bias towards the normal status's class and underperform on the abnormal status data, resulting in additional learning difficulty.}

To better understand the impact of limited labels and unbalanced training dataset on model performance, 
we use DeepSense~\cite{yao2017deepsense}, a popular multimodal model based on supervised learning, as the training model and the performance is tested during training. We utilize a public driver monitoring dataset, {Stressors}~\cite{taamneh2017multimodal}, for \needrev{stress status} detection where the training data is randomly reduced \nextrev{and the labeling rate is defined as the number of labelled samples divided by the total number of training samples.}

As illustrated in Fig.~\ref{fig:label_quantity}, with a balanced class distribution in training dataset, the accuracy drops by more than 15\% as the labelling rate decreases from 40\% to 5\% and the model converges more slowly with fewer labelled samples as well. Even if the labeling rate reaches 40\%, the model performance is still undesirable.
{Furthermore, Fig.~\ref{fig:label_class} shows that the accuracy, precision, and recall are evidently decreased with the unbalanced class distribution in training dataset and the training time is much longer compared to that with balanced training dataset.}
Therefore, how to harness the large amount of unlabeled data and limited well-labeled data in the unbalanced training dataset \nextrev{is essential for status identification.}


\vspace{-0.2cm}
\subsection{Data Recovery for Modality Missing} \label{sec:mtv_modality}
\begin{figure}[t]
\centering
\subfloat[Accuracy \label{fig:modality_missing_accuracy}]{
    \includegraphics[width=0.50\linewidth]{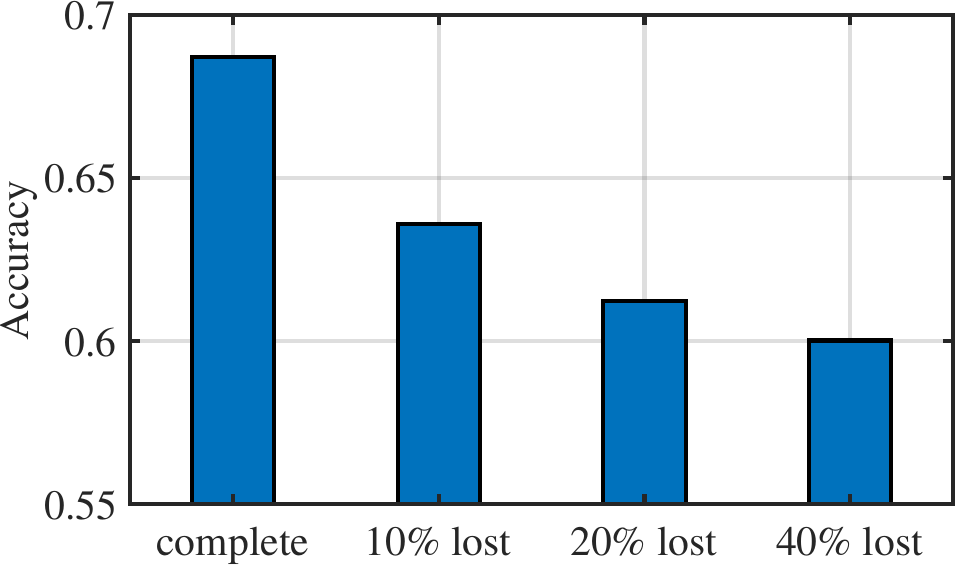}
}
\subfloat[Convergence \label{fig:modality_missing_convergence}]{
    \includegraphics[width=0.50\linewidth]{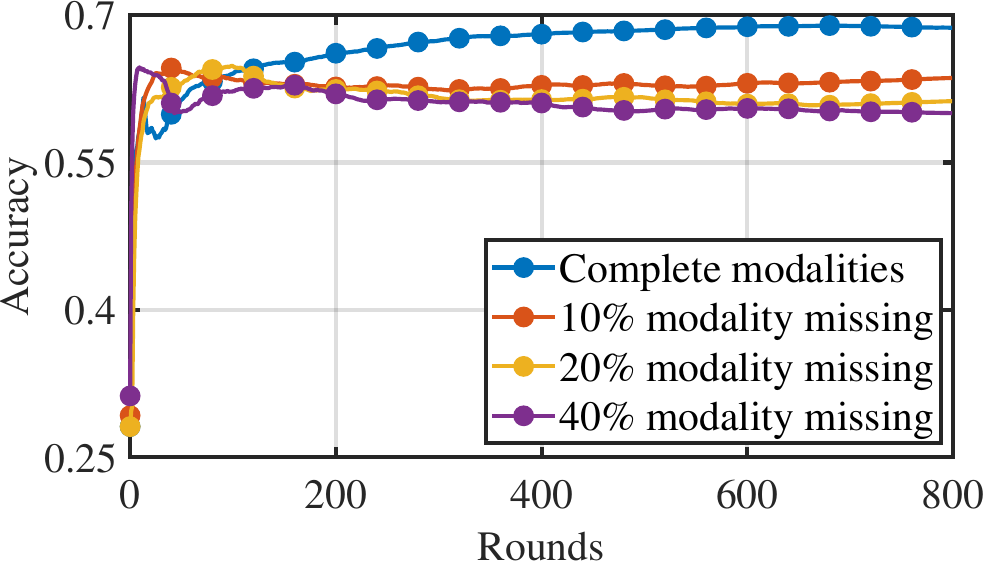}
}
    \caption{Performance comparison with heartbeat data missing in different proportions from the overall training dataset.}
    \label{fig:modality_missing_miss}
    \vspace{-2ex}
\end{figure}

A common assumption for multimodal learning approaches is that all modality data is available when feeding into the training model. However, it is not always applicable in real world due to environment and hardware limitations. Physiological signals may be lost due to sensor sensing limitations and visual data may be unavailable due to lighting and occlusion~\cite{ma2021smil}. 
Simply discarding samples or replacing missing modality data with zeros omits valuable contemporary information~\cite{suo2019metric}, thereby, such absence of modality data pose significant challenges on reliable status recognition, elevating the likelihood of incorrectly identifying status. 

We employ DeepSense model with concatenated feature to exhibit the testing accuracy for status detection where features are acquired from wearable sensors and a camera with complete modalities or with heartbeat data absence varying from 10\% to 40\%. \nextrev{Here the modality missing rate is the ratio of the zero-filling samples from the overall heartbeat samples in the training dataset.}
It is clear in Fig.~\ref{fig:modality_missing_miss} that the model trained with complete modalities outperforms those with information lost. Even with one modality available for training, the lack of auxiliary data significantly degrades the model performance, which underscores the vital role data recovery plays in status classification. 

\vspace{-0.2cm}
\subsection{Camera Rotation for Multi-object Monitoring} \label{sec:mtv_rotation}

\begin{figure}[t]
\centering
\subfloat[\hspace{2mm}\needrev{Comparison of a single camera \\ with fixed or adjusted angle.} \label{fig:camera_rotation_angle}]{
    \includegraphics[width=0.50\linewidth]{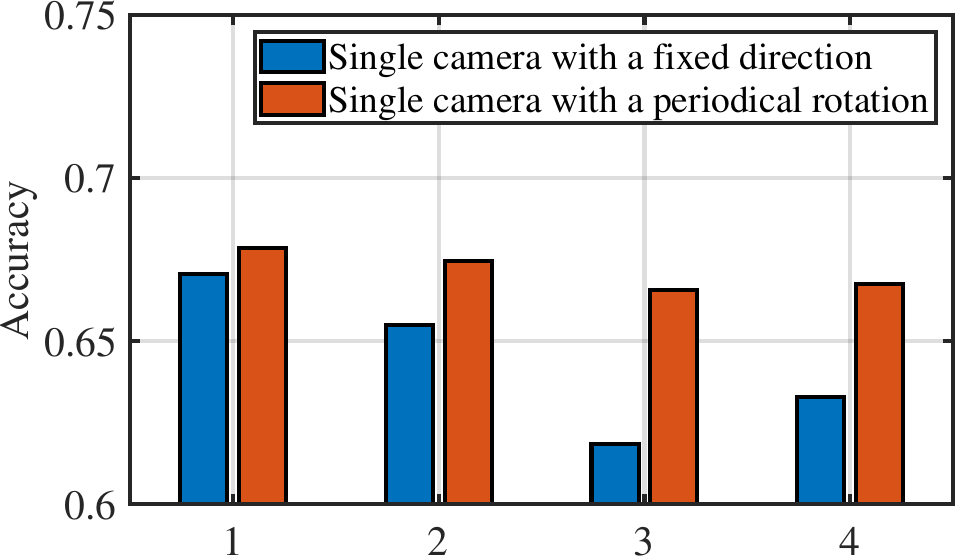}
}
\subfloat[Comparison of four fixed cameras \\ and a single camera rotation. \label{fig:camera_rotation_bandwidth}]{
    \includegraphics[width=0.50\linewidth]{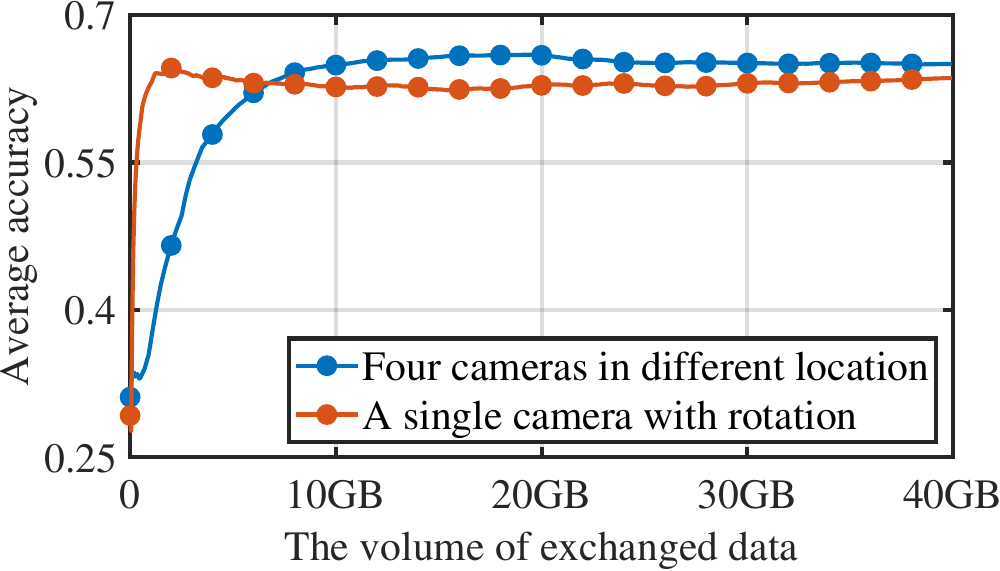}
}
    \caption{\newrev{Performance comparison with single camera rotation and other solutions for multi-object monitoring where 1,2,3, and 4 represent the driver, the passenger in assistant driver seat, and two passengers in back row.}}
    \label{fig:camera_rotation}
    \vspace{-2ex}
\end{figure}

\begin{figure}[t]
\centering
\subfloat[Modality missing \label{fig:camera_rotation_challenge_missing}]{
    \includegraphics[width=0.50\linewidth]{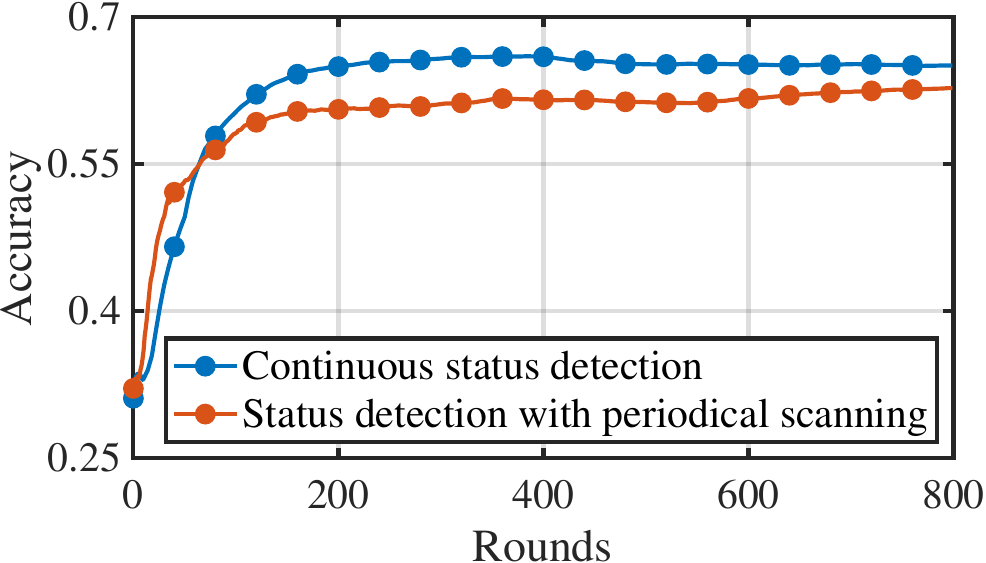}
}
\subfloat[Feature distribution \label{fig:camera_rotation_challenge_fusion}]{
    \includegraphics[width=0.50\linewidth]{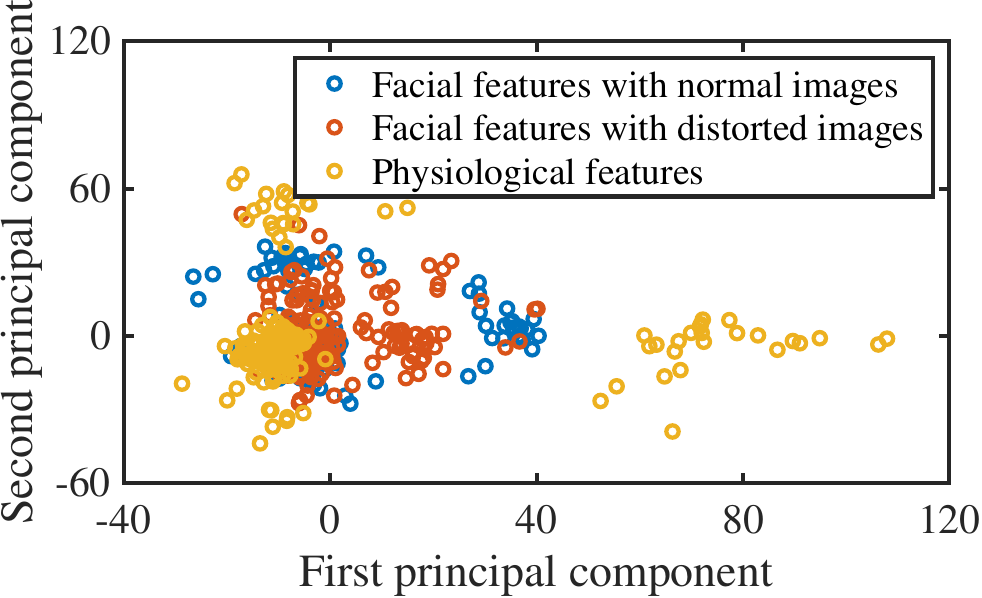}
}
    \caption{The impacts of rotating camera on status detection with distorted images and heartbeat signals.}
    \label{fig:camera_rotation_challenge}
    \vspace{-2ex}
\end{figure}


To monitor the abnormal status for both driver and passengers, it is essential to employ devices capable of tracking multiple objects simultaneously.
However, the typical configurations for in-car monitoring involves a camera mounted on the rear-view mirror to record facial information~\cite{jha2021multimodal, ortega2020dmd, angkititrakul2007utdrive}.
With this setup, not all passengers can be captured within the coverage of a single camera~\cite{jha2021multimodal} and the passengers in the back row may be obscured by front seats~\cite{ortega2020dmd}. 
{The limited coverage of the camera hinders continuous monitoring of passenger status, increasing the risk of missing critical events and potentially \rev{resulting in delayed detection which may cause serious consequences.}}

To evaluate the influence of the limited coverage and occlusion on status detection, \nextrev{we deploy different DeepSense models for each person with the facial information captured from the camera and physiological signals collected from wearable devices. 
The training performance is compared between a single fixed-direction camera and a periodically rotating camera for the simultaneous monitoring of driver and three passengers.}
Fig.~\ref{fig:camera_rotation_angle} shows that the testing accuracy significantly deteriorates for each passenger when using a fixed camera, as it is not capable of simultaneously capturing facial features for multiple passengers.
\nextrev{To address the limitations of the existing system for multi-object monitoring, it is necessary to seek a new alternative solution.}

In this regard, adjusting the angle of a single camera unlocks the potential for in-car monitoring of multiple objects, as a single camera with a periodic scanning from varying angles allows for an entire coverage of every passenger in both front and back seats.
We employ DeepSense in another experiment to compare model performance using a single rotating camera and four fixed cameras deployed in different locations for multi-object monitoring.
\nextrev{The average of the four testing accuracy from each person with the volume of the exchanged raw image data in Fig.~\ref{fig:camera_rotation_bandwidth} indicate that rotating camera achieves a considerable detection accuracy with much fewer data exchanges, confirming the potential of camera rotation as a promising solution.}

Despite the advantages of extended coverage and improved visibility under occlusion, turning the angle of camera for multi-object monitoring is still a non-trivial task.
We deploy DeepSense model to compare the performance of \rev{continuous monitoring against periodic capture with distorted images.}
{Fig.~\ref{fig:camera_rotation_challenge_missing} indicates that periodical rotation exacerbates modality missing when passengers are out of the field of view, thereby diminishing the auxiliary information crucial for recognition.
Moreover, \nextrev{we illustrate the first two components in the Principal Component Analysis (PCA) for the features extracted from camera and heartbeat samples.} As demonstrated in Fig.~\ref{fig:camera_rotation_challenge_fusion}, the 
motion blur and distortion by varying camera angles impedes discriminative feature extraction, thus \rev{enlarging the distribution gap} and adding complexity to align the distribution of distorted facial features with those of other modalities.
Therefore, it is imperative to design a more effective missing modality reconstruction for in-car multi-target monitoring.}

\vspace{-0.2cm}
\section{System Design} \label{sec:system_design}
In this section, we elaborate our \name framework, a new system for monitoring the status of both drivers and passengers with multimodal sensors in the car. Motivated by the insights from Sec.~\ref{sec:motivation}, {our key idea is to fully utilize the information from limited labels and large amount of unlabeled data through contrastive semi-supervised learning with adaptive threshold for class balance, while also reconstructing the missing modality to bridge the distribution gap so as to restore the complementary information for reliable status identification.} In the following, we first define our problem concretely, and then describe details of the system architecture.

\vspace{-0.2cm}
\subsection{Problem Statement and Overview} \label{sec:sd_overview}

\begin{figure}[t] 
\vspace{-2ex}
\centering
\includegraphics[width=\linewidth]{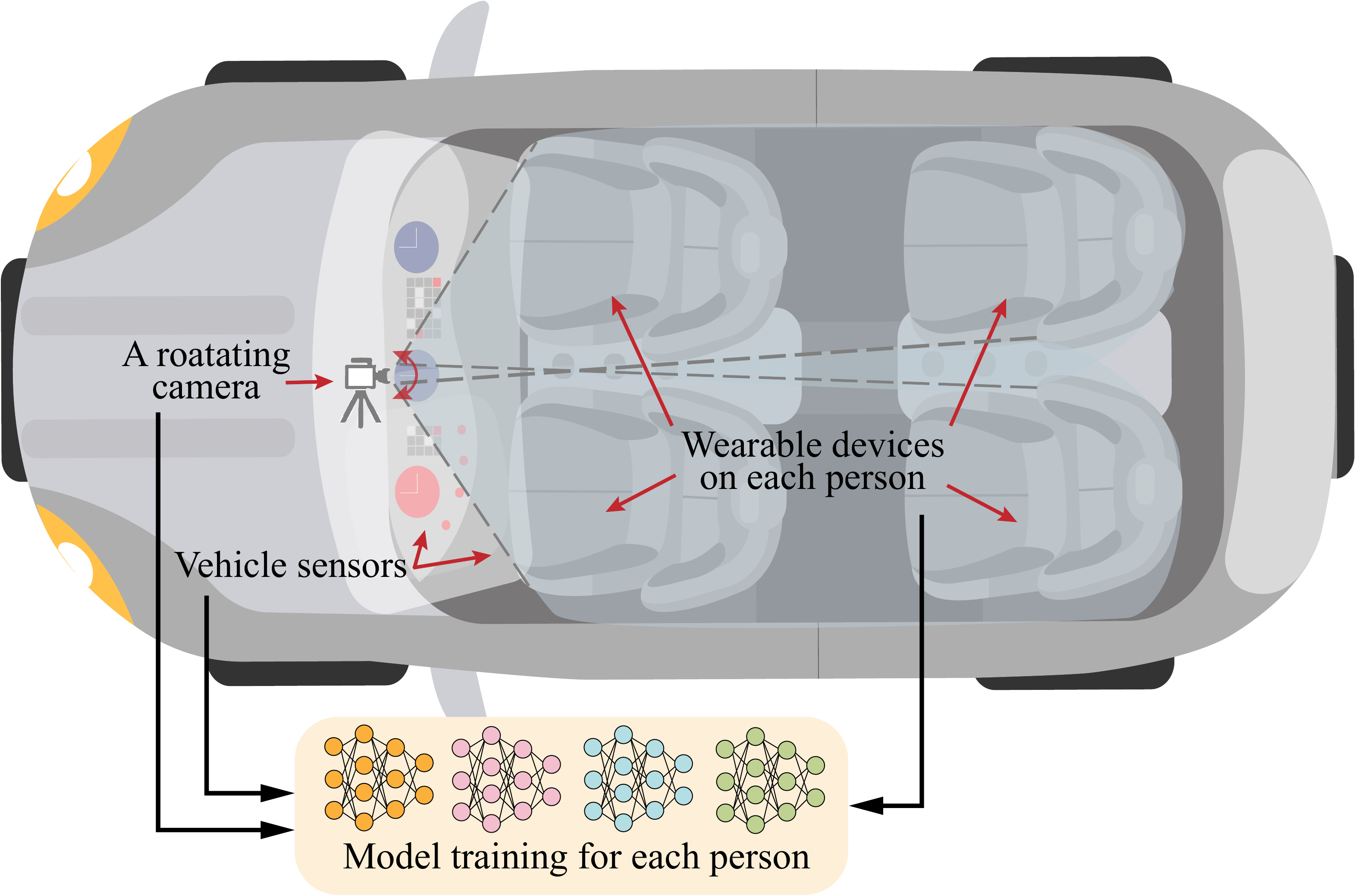}
\caption{The multimodal vehicle system of \name for abnormal status detection.}
\label{fig:scenario}
\vspace{-2ex}
\end{figure}

Our \name is designed to simultaneously monitor the status of both the driver and other passengers with continuously collected multimodal data, thereby enhancing not only driving safety to prevent accidents but also passenger safety to promptly respond to health emergency.
\nextrev{The abnormal statuses for the driver and passengers exhibit distinct facial and physiological patterns at the early stage, such as twisted facial expression, difficulty breathing, and irregular heartbeat~\cite{nvemcova2020multimodal, kaplan2015driver}. 
Therefore, as shown in Fig.~\ref{fig:scenario}, we utilize multiple sensors \needrev{(camera and wearable devices for the driver and passengers, and vehicle sensors only for the driver)} to capture these early signs in different perspectives and train models for each person, providing early identification and timely alarms of \rev{potential issues}.}
{Associating specific rotation angles with different persons allows passengers in different locations to be distinguished with a rotating camera.}


Although status detection has been extensively explored for in-car monitoring, the limitation of reliability and efficiency for informative representation learning emphasized in Sec.~\ref{sec:introduction} restrict the learning ability of the training model. 
{As illustrated in Sec.~\ref{sec:motivation}, the limited labeled data and class imbalance hamper training efficiency while modality missing leaves detection results unreliable}, forcing traditional algorithms to yield only comparable or even inferior performance. 
Especially, the use of rotating camera exacerbates the challenges by \rev{increasing the feature distribution gap and raising the possibility of low quality data and information lost.}

\begin{figure}[t] 
\centering
\includegraphics[width=\linewidth]{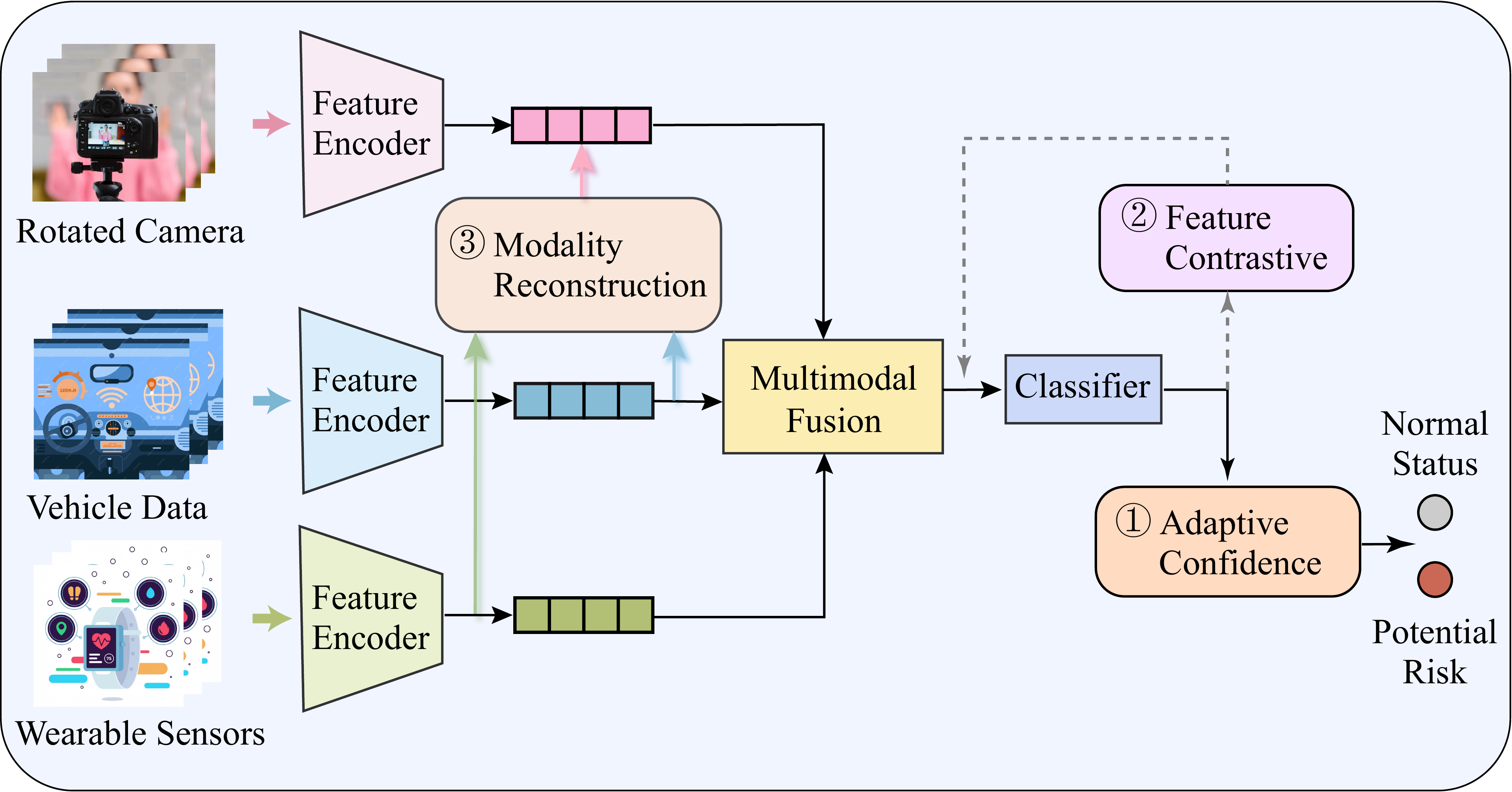}
\caption{The system overview of our \name.}
\label{fig:overview}
\vspace{-2ex}
\end{figure}
\nextrev{Our design consists of the following two crucial modules: adaptive threshold pseudo-labeling and missing modality reconstruction.
First, we design an adaptive threshold adjustment module to dynamically \rev{adjust the confidence threshold for each class, effectively} selecting high-confidence data to guide model training within imbalanced class distribution (Sec.~\ref{sec:adaptive_confidence}).
We also devise the contrastive loss to further leverage the rich information contained in low-confidence data, facilitating informative feature extraction with limited labels (Sec.~\ref{sec:contrastive_training}).
Second, we develop a modality reconstruction module to 
reconstruct the modality-specific features by analyzing the correlations between various modality distributions (Sec.~\ref{sec:distribute_approximate}) and transfer complementary information through mapping relationships (Sec.~\ref{sec:feature_mapping}) to achieve accurate recovery with limited labeled data (Sec.~\ref{sec:meta_training}). 
Fig.~\ref{fig:overview} shows the overall system architecture of our \name.}

\vspace{-0.2cm}
\subsection{Adaptive Threshold Pseudo-Labeling} \label{sec:confidence_training}


Recalling Sec.~\ref{sec:mtv_label}, labeled data is difficult to acquire in our \name because of the unreliable subjective evaluation. 
{A small amount of labeled data may fail to capture the full feature distribution, leading to poor model performance on unseen data.}
To address this, semi-supervised learning (SSL) has emerged as an effective approach which leverages abundant unlabeled data to enhance performance with limited labeled samples. 
{The core idea of SSL is to exploit valuable information in a large amount of unlabeled data by generating reliable labels for these unlabeled examples, effectively increasing the labeled dataset available for training.
This is achieved through two key techniques: pseudo-labeling~\cite{lee2013pseudo, mclachlan1975iterative}, which selects trustworthy labels to unlabeled data, and consistency regularization~\cite{bachman2014learning, raffel2020exploring}, which ensures that the model's predictions remain consistent between the weakly and strongly augmented versions of the same data.}

Specifically, given a multimodal dataset $\mathbf{X}$, we treat $x^m_i$ as $m$-th modality in the $i$-th multimodal sample, where $\mathbf{X}=\{x_1, x_2, ..., x_N\}$ contains $N$ multimodal data samples and each sample $x_i=\{x_i^m, m=1, ..., M\}$ consists of synchronized sensing data from $M$ modalities. SSL computes the prediction distribution $p(y_i|a(x_i)))$ from unlabeled data $x_i$ with weak feature augmentation $a(\cdot)$ (e.g., \needrev{the concatenation of features from different modalities}) to select pseudo labels. The excess of the highest prediction probability $\max(p(y_i|a(x_i)))$ above a confidence threshold $\tau$ indicates that the pseudo label is reliable as the unlabeled data provides informative features. Consequently, this unlabeled data $x_i$ with its pseudo label $\hat{y}_i = {\rm argmax}(p(y_i|a(x_i)))$ is added to the labeled dataset $D_l$ for subsequent model training as follows
\begin{equation} \label{eq:label_dataset}
D_l = D_l \cup \{x^{m}_i, \hat{y}_i, m=1,...,M\}
\end{equation}


{According to consistency regularization, the model should output similar predictions when feeding perturbed versions of the same data~\cite{sohn2020fixmatch}}. To this end, the cross-entropy function is enforced to the selected pseudo label $\hat{y}_i$ and the model prediction $p(y_i|A(x_i))$ with a strongly-augmented version $A(\cdot)$ of $x_i$ (e.g., \needrev{adding noise for physiological data and distortion for facial information}), expecting the model to output consistent results. Therefore, the unsupervised loss for high-confidence unlabeled data can be represented as
\begin{equation} \label{eq:consistency_loss}
L_{pl} = \sum_{i=1}^{N_u} {\mathbf{1}(\max(p(y_i|A(x_i))) \ge \tau)) H(\hat{y}_i, p(y_i|A(x_i))))}
\end{equation}
where $N_u$ is the number of samples in the unlabeled dataset and $\tau$ represents a predefined confidence threshold that controls the quality and quantity of pseudo labels generated from unlabeled data.

Meanwhile, for labeled data, the model predicts labels from strongly augmented features and updates parameters using the standard cross-entropy loss $L_{cls}$.
\begin{equation} \label{eq:classification_loss}
L_{cls} = \sum_{i=1}^{N_l} {H(y_i, p(y_i|A(x_i))))}
\end{equation}
where $N_l = N - N_u$ is the number of labels. Finally, with the balance weight $\lambda$ for unsupervised loss $L_{pl}$, the overall training loss $L_{ssl}$ is typically as
\begin{equation}
L_{ssl} = L_{cls} + \lambda L_{pl}
\end{equation}



\begin{figure}[t] 
\centering
\includegraphics[width=\linewidth]{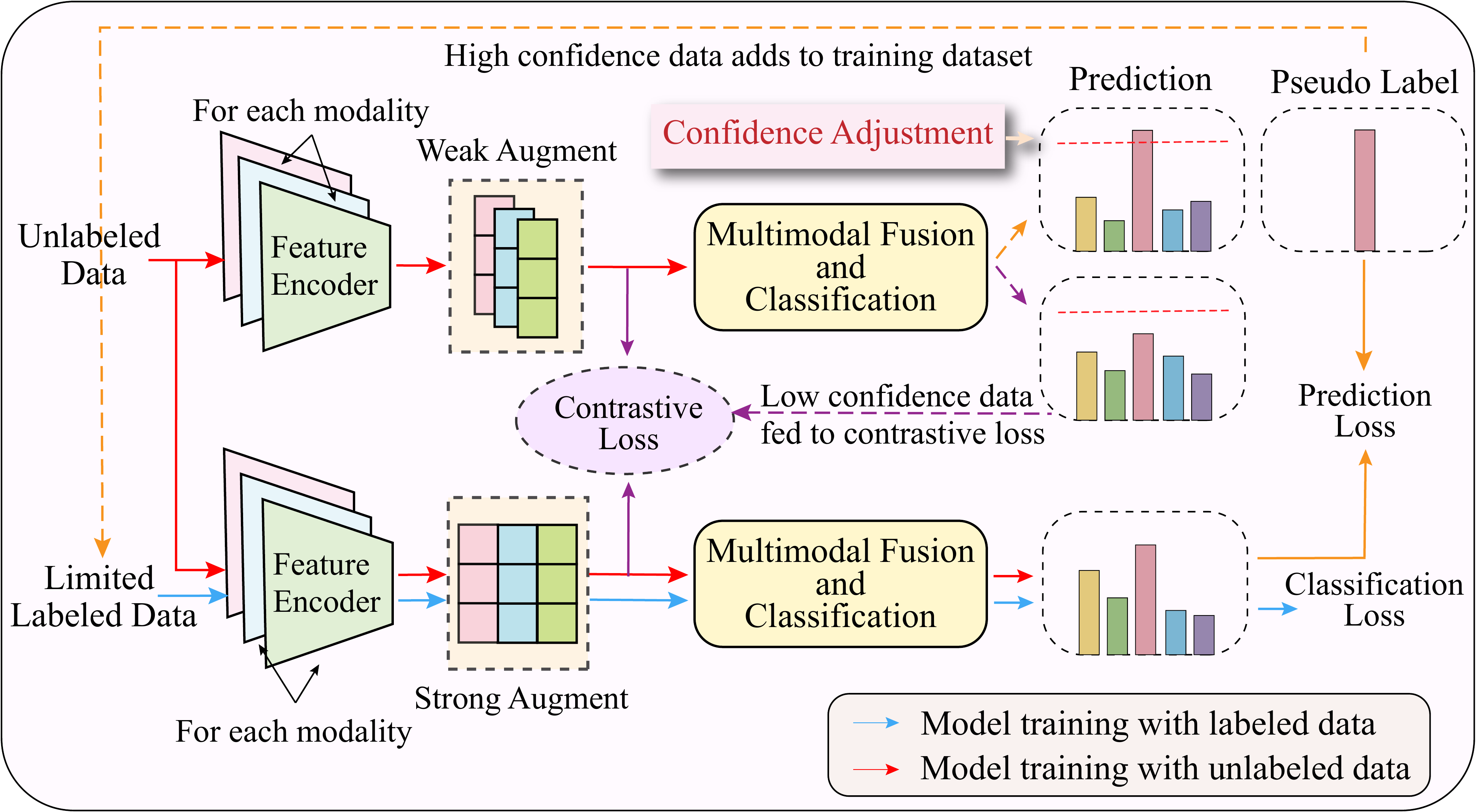}
\caption{Confidence based multimodal training with adaptive threshold adjustment.}
\label{fig:confidence_training}
\vspace{-2ex}
\end{figure}

Recalling in Sec.~\ref{sec:mtv_label} that the scarcity of data from abnormal status leads to an imbalanced class distribution.
Training model with class imbalance causes model to prioritize the detection normal states while neglecting features from the class of abnormal status, making it difficult to learn informative features for abnormal status.
However, in high-risk applications such as \newrev{medical diagnostics}, the inability to detect abnormal status promptly is particularly dangerous and even endangers lives, emphasizing the necessity of training models to identify abnormal status despite class imbalance.
To address this issue, as illustrated in Fig.~\ref{fig:confidence_training}, we propose an adaptive adjustment of the confidence threshold in semi-supervised learning, which generates high-confidence, class-balanced pseudo labels for model training.

Moreover, to prevent involving incorrect model predictions for pseudo labels, modern SSL algorithms~\cite{berthelot2019mixmatch, sohn2020fixmatch, xie2020unsupervised} only leverage unlabeled data with artificial labels above a predefined confidence threshold (typically selected as 0.95) for model training while those below threshold {remain as low-confidence unlabeled data}. However, excluding low-confidence unlabeled samples in model update {restricts the efficiency of informative feature extraction, requiring a large training rounds to reach competitive results.}
Therefore, we further explore a confidence-based training scheme that fully utilizes the unlabeled data with low confidence, facilitating feature extraction with the rich information in those low-confidence unlabeled data.

\subsubsection{\textbf{Adaptive threshold adjustment}} \label{sec:adaptive_confidence}

In semi-supervised learning, the threshold is crucial for selecting high-confidence pseudo labels.
However, a fixed confidence threshold is used for pseudo-labeling in traditional semi-supervised learning algorithms~\cite{berthelot2019mixmatch, sohn2020fixmatch, xie2020unsupervised}, which excludes a significant portion of the unlabeled data from model training at the early stage when the class confidence is generally low, hindering model ability to learn informative feature representations.
Recent works~\cite{zhang2021flexmatch, guo2022class} have addressed this issue by dynamically adjusting the threshold for each class based on the number of pseudo labels, thus adapting to various training stages and enhancing model performance across extensive semi-supervised learning benchmarks. 
In practice, both the unlabeled and labeled data suffer from a severe class imbalance due to the data scarcity in abnormal status, making it inadequate to adjust the threshold solely based on the number of pseudo labels generated from the unlabeled data.
Even if the selected pseudo labels are class-balanced, the total label quantities among various classes remain imbalanced when combining the original labels with the selected pseudo labels.

Therefore, we design an adaptive threshold adjustment scheme for pseudo-labeling with balanced class distribution and quantity. 
Our \name aligns the class quantity by adjusting the threshold for each class according to the total number of labeled data and pseudo labels selected from the unlabeled data, thereby mitigating the impact of class imbalance on model training.

More specifically, at the $t$-th training round, the number of labels for class $c$ is given by $\sigma_t^l(c)=\sum_{i=1}^{N_l} \mathbf{1}(y_i=c)$ and the number of pseudo labels selected for class $c$ is calculated as
\begin{align} \label{eq:learning_status}
\begin{split}
\sigma_t^u(c) = \sum_{i=1}^{N_u} [ \mathbf{1}(\max(p(y_i|a(x_i)) \ge \tau_t(c))) \\
\cdot \mathbf{1}({\rm argmax}(p(y_i|a(x_i))=c) ]
\end{split}
\end{align}
where $\tau_t(c)$ is the adaptive threshold for class $c$ and $\mathbf{1}(\cdot)$ represents the indicator function. After one round training, the total number of labels in class $c$ is calculated as
\begin{equation} \label{eq:number_labels}
\gamma_t(c)= {\sigma_t^l(c)} + {\sigma_t^u(c)}
\end{equation} 

Then, we calculate the proportion of the label quantity in class $c$ as its cumulative distribution to \rev{reflect the class distribution imbalance}. 
Define the total number of classes as $C$, the cumulative distribution of class $c$ is represented as
\begin{equation} \label{eq:cumulative_distribution}
p_t(c)= \frac{\gamma_t(c)}{\sum_{c=1}^{C} \gamma_t(c)}
\end{equation}

To achieve a balanced class distribution during training, we utilize KL divergence as a guidance for threshold adjustment. 
By quantifying the deviation between the current cumulative distribution and the target class-balanced distribution, the model adjusts thresholds to select varying numbers of samples for different classes, encouraging a closer class quantity across classes.
For the current empirical distribution of pseudo labels \needrev{$P(c)=p_t(c)$} and the class-balanced target distribution \needrev{$Q(c)=1/C$}, the KL divergence is defined as
\begin{equation} \label{eq:KL_divergence}
D_{KL}(P||Q) = \sum_c^C {P(c) \log{\frac{P(c)}{Q(c)}}}
\end{equation}

Lastly, the adaptive pseudo-labeling threshold for class $c$ is
\begin{equation} \label{eq:threshold_adjust}
\tau_t(c) = \min\{(p_t(c) + \tau - D_{KL}(p_t)), \tau_h\}
\end{equation}
where $\tau$ is the predefined base threshold (e.g., 0.95) and $\tau_h$ is the upper bound of threshold (usually set to 0.95) which avoids threshold saturation (i.e., exceed 1.0) to leave out most of potentially useful data for training.

Importantly, KL divergence is crucial for balancing data distribution, with high KL divergence reflecting large class distribution differences. The threshold $\tau_t(c)$ for minority class is appropriately lowered to generate more pseudo labels, while a higher $p_t(c)$ for majority classes ensures their thresholds are less influenced by high KL divergence. This approach facilitates a gradual alignment of the training class distribution with the target class-balanced distribution, thereby promoting a more balanced class quantity in the next training round.

\subsubsection{\textbf{Semi-supervised training with contrastive learning}} \label{sec:contrastive_training}
Recalling Sec.~\ref{sec:adaptive_confidence}, high-confidence unlabeled data \rev{(with the highest prediction probability exceeding the threshold)} are selected as pseudo labels for training with more balanced class distribution through dynamic threshold adjustment, allowing model to better extract informative features for each class.  
However, the remaining low-confidence unlabeled data, despite less distinctive, still contain valuable features that contribute additional information to improve status classification.

Current approaches~\cite{berthelot2019mixmatch, sohn2020fixmatch, xie2020unsupervised, zhang2021flexmatch, guo2022class} only rely on high-confidence data for training. However, in the early training stage, 
{limited high-confidence pseudo labels typically represent only easily classifiable samples, failing to capture the full data distribution and resulting in model bias towards well-represented classes.}
In contrast, incorporating low-confidence unlabeled data in model training provides more diverse training samples, encouraging the model to explore hard-to-classify features and gradually improve performance. 
Therefore, including low-confidence data in training is essential to further enhance feature extraction with limited labels.

Labels for low-confidence unlabeled data remain undetermined due to the unreliable model predictions. Fortunately, contrastive learning, an emerging unsupervised learning approach that does not require labels, can effectively learn effective feature representations 
by pulling the features of unlabeled samples from the same latent class together and pushing the features of different classes apart. 
\nextrev{Incorporating contrastive learning into model significantly enhance data diversity by leveraging the large amount unlabeled data for discriminative feature extraction, thereby facilitating the feature extraction from hard-to-classify samples.}

More specifically, we treat pairs of the augmented multimodal features $\{ a(x_i), A(x_i) \}$ from the same sample $x_i$ as the positives $p_i^+$, versus pairs of the multimodal features from different samples $\{ a(x_j), A(x_j) \}$ ($j \ne i, j \in \{1, ..., N\}$) as negatives $p_j^-$. By minimizing the contrastive loss formulated below, the positive features are brought together while the negative features are separated apart for data without labels.
\begin{equation}  \label{eq:contrastive_loss}
L_{con} = - {\log\frac{{d(\{{p_i}^+\})}}{d(\{{p_i}^+\})+{\sum_{s=1}^{S}{d(\{{p_k}^-\})}}}}
\end{equation}
where {$S$ is the number of negative samples (e.g., the batch size)} and $d(\cdot)$ is the cosine similarity adjusted by a hyper-parameter $T$ (\nextrev{typically 0.05}), which can be formulated as
\begin{equation}
d(\{{p_i}\}) = \exp \left( \frac{a(x_i) \cdot A(x_i)}{||a(x_i)|| \cdot ||A(x_i)||} \cdot \frac{1}{T} \right)
\end{equation}

With the contrastive loss $L_{con}$ to learn discriminative feature representations, our \name fully utilizes low-confidence model prediction ($\max(p(y_i|a(x_i))) < \tau$) to boost the model performance.
To this end, by combining the contribution of labeled data and high-confidence unlabeled data to the model, the overall training loss for status detection is defined as
\begin{equation} \label{eq:ssl_loss}
L_{ssl} = L_{cls} + \lambda_p L_{pl} + \lambda_c L_{con}
\end{equation}
where $\lambda_p$ and $\lambda_c$ are weight parameters that {control the contributions of unlabeled data in model training.}
\vspace{-0.2cm}
\subsection{Missing Modality Reconstruction} \label{sec:modality_reconstruction}

\begin{figure}[t]
\centering
\subfloat[Feature distribution \label{fig:modality_missing_distribution}]{
    \includegraphics[width=0.465\linewidth]{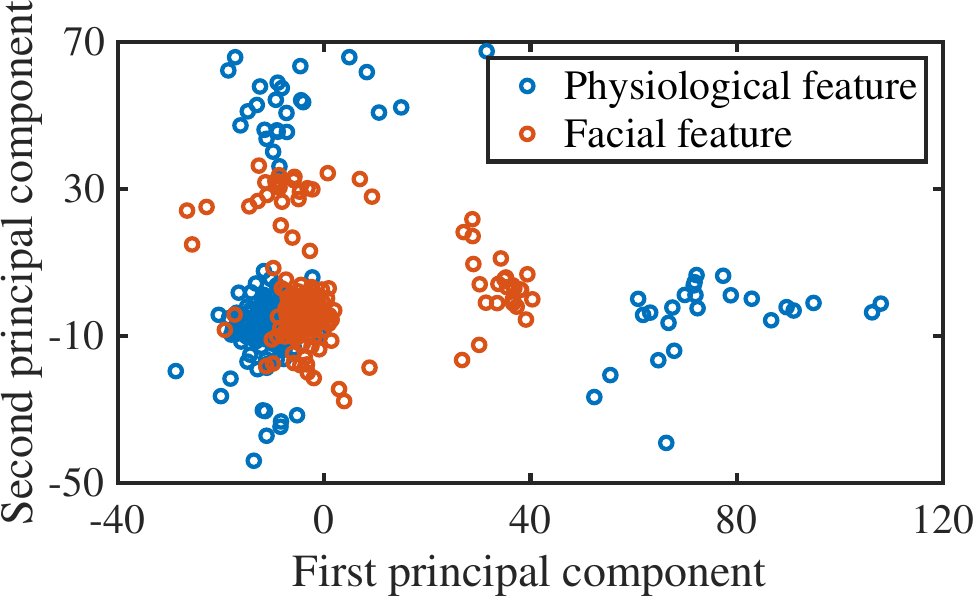}
}
\subfloat[Heartbeat recovery \label{fig:modality_missing_predict}]{
    \includegraphics[width=0.465\linewidth]{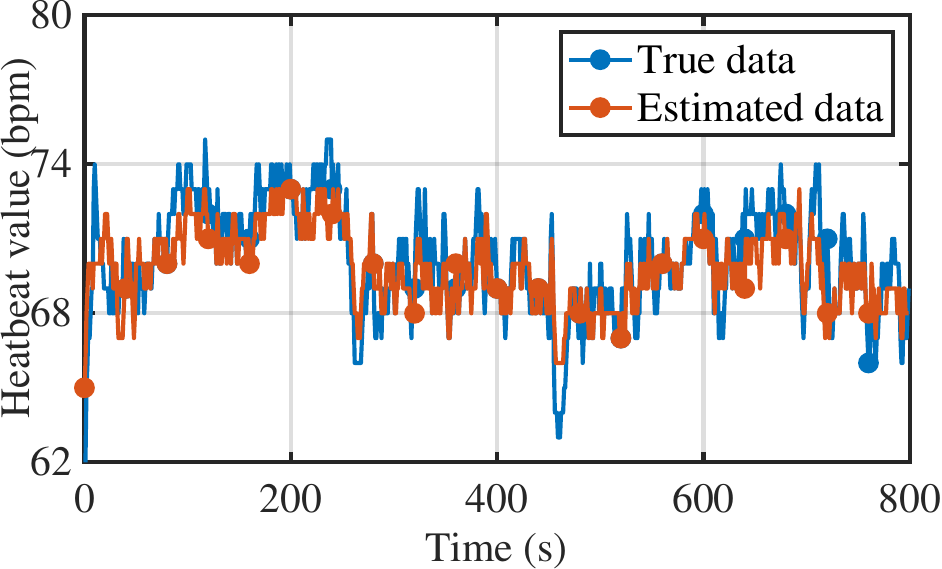}
}
    \caption{Impact of feature distribution gap on the missing heartbeat data recovery with available camera data.}
    \label{fig:modality_missing_impute}
    \vspace{-2ex}
\end{figure}

Recalling to Sec.~\ref{sec:mtv_modality} that a vehicle system cannot always access full-modality data, we propose a modality reconstruction network to recover the feature representations of the missing modality.
Multimodal data shares similar semantic meanings from different perspectives, \newrev{hence through learning the hidden relationship between modalities, the missing modality can be reconstructed according to} other available modalities~\cite{suo2019metric}.
However, existing methods usually recover the missing data directly across the common feature extraction~\cite{zhao2021missing, zhou2017anomaly, zhou2021memorizing, perera2019ocgan}, which {overlooks different complementary information provided by each modality, resulting in distribution gap that prevents the accurate recovery of missing data.}
As mentioned in Sec.~\ref{sec:mtv_rotation}, {severe missing modality and image distortion due to camera rotation reduce the shared information between modalities}, highlighting the importance of complementary information for data recovery in our scenario.

\nextrev{To motivate our design of modality reconstruction module}, we deploy MMIN~\cite{zhao2021missing}, one of the state-of-the-art multimodal networks for handling missing modalities, to learn the missing heartbeat representations from camera samples in the {Stressors} dataset and then test the recovery performance on heartbeat samples. We apply PCA to the features extracted from heartbeat and camera samples and visualize the first two PCA components to illustrate their feature distribution.
Fig.~\ref{fig:modality_missing_distribution} demonstrates that there is an evident gap in the feature distribution of different modalities and Fig.~\ref{fig:modality_missing_predict} reveals that only relying on the common information in available modalities results in inconsistent recovery between the recovered and true data.
\nextrev{Since the distribution gap impedes the direct estimation of the missing data, it is imperative to propose a new data imputation technique to handle this issue.}

{Therefore, instead of directly generating the missing data based on the shared features, 
we reconstruct distribution of missing modality to restore the modality-specific feature representation with the correlations between modality distributions.
By recovering data based on the cross-modality feature transformation from available data}, the generated data contains the shared information among modalities and the complementary information of the missing modality as well, hence mitigating the distribution gap between the recovered and the true data.

We build a reconstruction module to learn the feature mapping function $g_{\phi}(\cdot)$ for distribution transferring. 
\nextrev{Without loss of generality, below we only demonstrate the reconstruction of the modality $m$ from other available modalities $m'$ ($m'\ne m, m'\in \left\{ 1, ..., M \right\}$), where the feature representations $\mathbf{Z}^{{m}'}$ are extracted by the feature encoder from modality $m'$ independently.}
{To maximize the mutual information that ensures informative feature relationships to be transferred from available modalities $\mathbf{Z}^{{m}'}$ to the missing modality ${z}^{m}$}, the objective is to minimize the conditional entropy $H(\mathbf{Z}_{}^{m}|\mathbf{Z}_{}^{{m}'})$ in the feature space across the data distribution. Higher entropy implies greater uncertainty in data recovery, thus the reconstruction objective is denoted as
\begin{equation} \label{eq:conditional_entropy}
\min H(\mathbf{Z}_{}^{m}|\mathbf{Z}_{}^{{m}'}) =  \min {E_{P(\mathbf{Z}_{}^{m}, \mathbf{Z}_{}^{{m}'})} (-\log{P(\mathbf{Z}_{}^{m}|\mathbf{Z}_{}^{{m}'})})}
\end{equation}

\begin{figure}[t] 
\centering
\includegraphics[width=\linewidth]{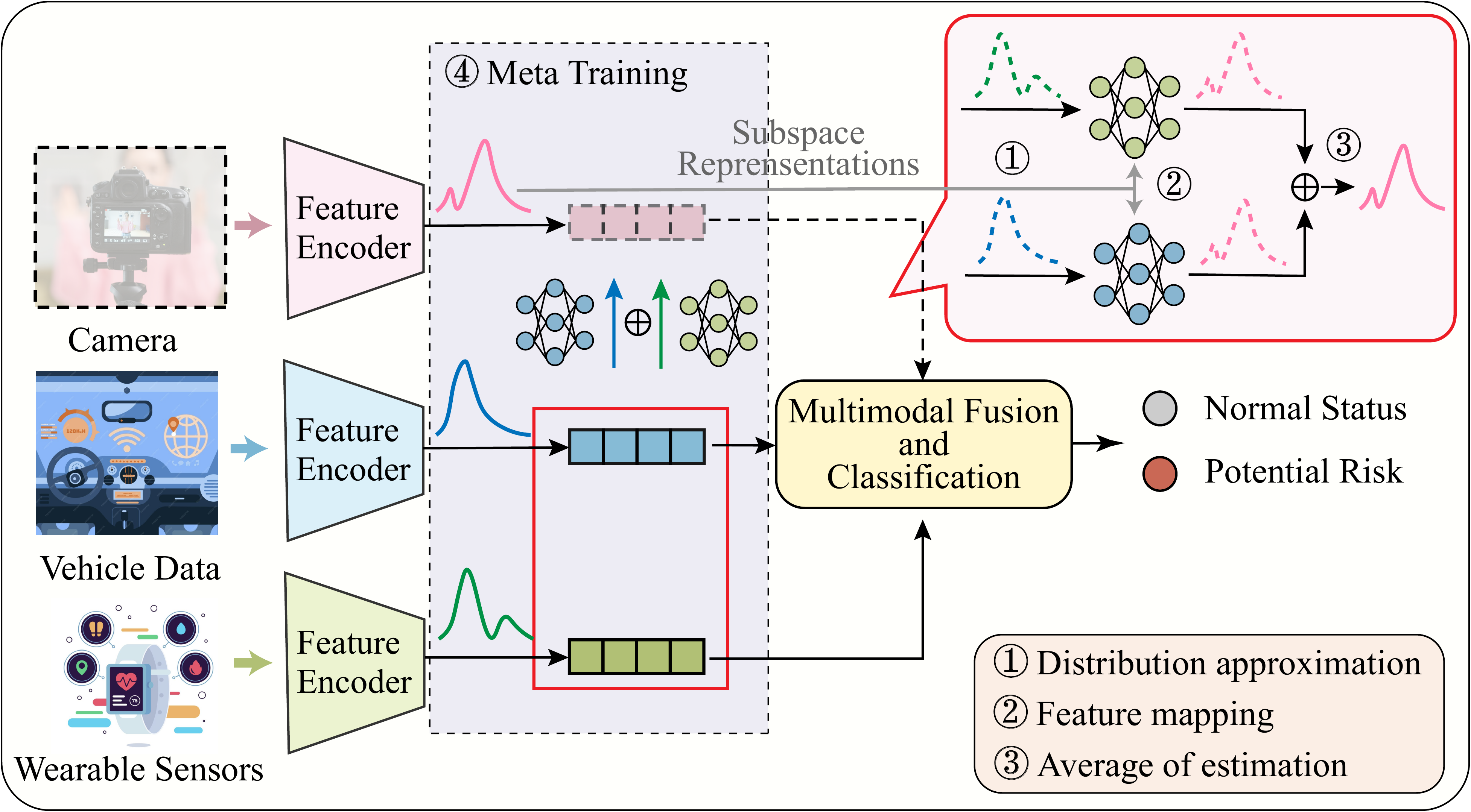}
\caption{Modality reconstruction module with meta learning.}
\label{fig:distribution_reconstruction}
\vspace{-2ex}
\end{figure}

As shown in Fig.~\ref{fig:distribution_reconstruction}, we first estimate the {informative structure representations} of the missing modality and other available modalities to preserve their modality-specific characteristics, and then train a feature mapping network to reconstruct the missing modality based on {approximated distribution}. Finally, the modality reconstruction is achieved by recovering the feature representations of missing samples with the averaged combination of distribution transferring from multiple modalities. 
{The model strives to maintain key features of the missing modality by leveraging cross-modality correlations from the distribution mapping, thus ensuring that the recovered data retains its modality-specific features.}

\subsubsection{\textbf{Distribution approximation}} \label{sec:distribute_approximate}
The extensively studied paradigm for data recovery~\cite{zhao2021missing, zhou2017anomaly, zhou2021memorizing, perera2019ocgan} focuses on directly recovering missing data from available modalities by discovering the hidden relationship between modalities through common feature extraction. 
However, shared information between modalities cannot fully encompass the distinct complementary information of each modality since different modalities exhibit inconsistent feature distributions due to their distinctive characteristics as shown in Fig.~\ref{fig:modality_missing_impute}. 
This means that the modality-specific feature representations cannot be reconstructed when simply relying on shared information for recovery, leading to inaccurate recovered data. 
To compensate for the distribution gap caused by modality-specific information, we utilize the limited data samples from the missing modality {in the labeled dataset} to exploit its distribution characteristics. \nextrev{By analyzing the correlations between various modality distributions, we reconstruct the characteristics of missing modality from the restored distribution, thereby enabling accurate data recovery.}

{The redundant information in high-dimensional sensor data makes it hard to visualize similarities between data samples, complicating the task of capturing the intrinsic structure of the data distribution.} Fortunately, many research efforts reveals that data distributions of images and time-series signals often lie in low-rank subspaces~\cite{liu2011latent, rao2009motion, ma2007segmentation}, which retain essential information while eliminating redundancy, providing a more concise representation of the data structure.
Therefore, constructing a low-rank subspace representation of modality distributions is pivotal for effectively extracting informative modality characteristics.

PCA~\cite{labrin2020principal}, a well-known technology for dimension reduction, selects principal components by identifying the directions with higher singular values in the data space, ensuring that the most important characteristics of the original data distribution are retained. 
Inspired by this observation, our \name utilizes {limited labeled data from the missing modality} to approximate its distribution through PCA for low-rank subspace construction.
This subspace representation preserves key distribution characteristics, serving as a guidance for subsequent feature mapping and data recovery.

Denote the feature representations of the $m$-th modality after the feature encoder as $\mathbf{Z}^{m}=\{z_i^m, i=1, ..., N\}$ where $\mathbf{Z}^{m} \in \mathbf{R}^{N \times F}$ with the feature length of $F$, we perform PCA on all $N$ samples in $\mathbf{Z}^{m}$ to extract the distribution structure of the missing modality. 
The approximated distribution $\mathbf{P}^{m}$ projected on the low-rank subspace contains the main characteristics of modality distribution, which is represented as $\mathbf{P}^{m}=\mathbf{Z}^{m} \mathbf{V}^{m}$ where $\mathbf{V}^{m} \in \mathbf{R}^{F \times K}$ ($K \ll N$) is the principal component matrices, representing the low-rank subspace of modality $m$. 
We select the first $K$ principal components (e.g. $K=4$), keeping the simplified yet effective representation for distribution approximation. 

\subsubsection{\textbf{Feature mapping}} \label{sec:feature_mapping}
The complementary information specific to each modality causes the distribution gap between modalities, challenging data recovery with only shared information.
\nextrev{Camera rotation further exacerbates the feature distribution differences between visual and physiological modalities, intensifying the challenge of data recovery through common features alone.}
To address this issue, we design a feature mapping network that learns the relationships between the distributions of different modalities.
Instead of recovering missing data from common information between modalities, our \name transfers complementary information through mapping relationships, allowing for the reconstruction of missing modality characteristics from available modalities. 
By estimating the distribution of missing modality, the modality-specific features could be restored in recovered data, thereby mitigating the distribution gap and improving recovery accuracy.

Despite the distinct features emphasized by each modality leading to inconsistent distributions, different modalities share similar semantics as \rev{they capture various aspects of the same event within a common context}. This semantic similarity unleashes the potential for distribution transformation between modalities by aligning their semantic representations. 
Specifically, using the distribution approximations of various modalities obtained through PCA, we learn a feature mapping function $\mathbf{F}^{m'}$ that maps the distributions of available modalities $\mathbf{P}^{m'}$ to the missing modality $\mathbf{P}^{m}$ based on semantic similarity, expressed as $\mathbf{P}^{m}=\mathbf{P}^{m'} \mathbf{F}^{m'}$. This distribution mapping transfers the complementary information across feature distributions from different modalities, thereby bridging the distribution gaps between them.

To recover missing data, we first reconstruct the distribution of missing modality through feature mapping to restore the modality-specific feature representation, and then estimate the original data based on its low-rank subspace.
Since the principal components $\mathbf{V}^{m}$ allows the projection of approximated distribution $\mathbf{P}^{m}$ back to the original data distribution $\mathbf{Z}^{m}$ through $\mathbf{Z}^{m}=\mathbf{P}^{m}{(\mathbf{V}^{m})}^{T}$, the reconstruction of the missing modality $\mathbf{\hat{Z}}^{m}$ can be estimated as follows:
\begin{equation}
\mathbf{\hat{Z}}^{m} = (\mathbf{P}^{m'} \mathbf{F}^{m'}){(\mathbf{V}^{m})}^{T} = \mathbf{Z}^{m'}(\mathbf{V}^{m'} \mathbf{F}^{m}){(\mathbf{V}^{m})}^{T} 
\end{equation}

{When multiple modalities are available for feature mapping, our \name averages the estimated distributions from each modality.} By integrating information from all modalities, this averaging process helps smooth out outliers and biases present in distribution transformations, hence providing a more stable and reliable estimation. 
We implement the mapping relationship $\mathbf{F}^{m'}$ as a neural network with trainable parameters $\mathbf{W}^{m}$, thus for each missing data sample $x_{i}^{m}$, its recovered feature representation $z_{i}^{m}$ can be expressed as the average of feature mappings from other modalities $z_{i}^{m'}$, which is given by
\begin{equation}
\hat{z}_{i}^{m} = \frac{1}{M-1} \left[ \sum_{m'} {z_{i}^{m'}} \cdot \mathbf{W}^{m} \right] \cdot {(\mathbf{V}^{m})}^{T}
\end{equation}
where $\mathbf{W}^{m}$ is 
used to convert the feature representations $z_{i}^{m'}$ of other modalities into the feature representations $z_{i}^{m}$ of missing modality. 

{Since the low-rank subspace $V^{m}$ of the missing modality retains its distribution characteristics}, \nextrev{the recovered data can align with missing modality distribution}, overcoming the loss of modality-specific information due to \nextrev{distribution inconsistency among modalities} and thus accurately restoring the data.
The recovered data $\hat{z}_{i}^{m}$, along with other available data $z_{i}^{m'}$, are then fed to the adaptive threshold pseudo-labeling module in Sec.~\ref{sec:confidence_training} for further training to identify abnormal status with the semi-supervised learning framework.



\subsubsection{\textbf{Objective function}} \label{sec:missing_objective}
Modality reconstruction aims to approximate the distribution of the missing modality from other modalities and accurately recover the data based on the distribution characteristics of the missing modality. 
To evaluate the performance of modality reconstruction, we leverage the uncertainty of recovering missing data $\mathbf{\hat{Z}}_{}^{m}$ from other available data $\mathbf{Z}_{}^{{m}'}$ {to measure the reliability of the distribution transformation, which is denoted as the conditional entropy $H(\mathbf{\hat{Z}}_{}^{m}|\mathbf{Z}_{}^{{m}'})$ in Eqn.~\eqref{eq:conditional_entropy}.}
Moreover, it is worth noting that the primary goal of in-car status monitoring is to achieve the accurate abnormal status identification where modality reconstruction is designed to provide auxiliary information from the recovered data for status detection.
Therefore, we account for the contribution of the reconstructed modality in the final status recognition, and the modality reconstruction loss for missing modality $m$ is defined as follows.
\begin{equation}
L_{recover} = H(\mathbf{\hat{Z}}_{}^{m}|\mathbf{Z}_{}^{{m}'}) + \lambda_r H(p({\mathbf{\hat{y}}}|\mathbf{\hat{Z}}_{}^{m},\mathbf{Z}_{}^{{m}'}), p({\mathbf{y}}|\mathbf{Z}_{}^{m},\mathbf{Z}_{}^{{m}'}))
\end{equation}
where $\lambda_r$ is the {balanced parameter that controls the contribution of modality reconstruction to detection accuracy} and $H(p({\mathbf{\hat{y}}}|\mathbf{\hat{Z}}_{}^{m},\mathbf{Z}_{}^{{m}'}), p({\mathbf{y}}|\mathbf{Z}_{}^{m},\mathbf{Z}_{}^{{m}'}))$ represents the cross-entropy loss between the model output $p({\mathbf{\hat{y}}}|\mathbf{\hat{Z}}_{}^{m},\mathbf{Z}_{}^{{m}'})$, which is based on the recovered data  $\mathbf{\hat{Z}}_{}^{m}$ and the reference $p({\mathbf{y}}|\mathbf{Z}_{}^{m},\mathbf{Z}_{}^{{m}'})$ from the true data $\mathbf{Z}_{}^{m}$.

Optimizing the conditional entropy $H(\mathbf{\hat{Z}}_{}^{m}|\mathbf{Z}_{}^{{m}'}) = E_{P(\mathbf{\hat{Z}}_{}^{m}, \mathbf{Z}_{}^{{m}'})} (-\log{P(\mathbf{\hat{Z}}_{}^{m}|\mathbf{Z}_{}^{{m}'})})$ involves \rev{calculating} the true distribution $P(\mathbf{\hat{Z}}_{}^{m}, \mathbf{Z}_{}^{{m}'})$, which is intractable~\cite{ma2021smil}.
Instead, we approximate it by introducing a variational distribution $Q(\mathbf{\hat{Z}}_{}^{m}|\mathbf{Z}_{}^{{m}'})$, which provides an easily optimized lower bound
\begin{equation}
{E_{P(\mathbf{\hat{Z}}_{}^{m}, \mathbf{Z}_{}^{{m}'})} (-\log{Q(\mathbf{\hat{Z}}_{}^{m}|\mathbf{Z}_{}^{{m}'})})}
\end{equation}

Since $H(P|Q)=E_{P}(-\log Q)+D_{KL}(P|Q)$, where the KL divergence $D_{KL}(P|Q)$ is non-negative, maximizing this lower bound indirectly minimizes the KL divergence. This process makes the variational distribution $Q$ closer to the true distribution $P$, thereby optimizing the conditional entropy $H(P|Q)$.
Here $Q(\mathbf{\hat{Z}}_{}^{m}|\mathbf{Z}_{}^{{m}'})$ can be modeled as a Gaussian distribution like $N(\mathbf{Z}^{m}|\mathbf{\hat{Z}}_{}^{m}, \sigma \mathbf{I})$~\cite{goodfellow2014generative}. By ignoring the constants derived from Gaussian distribution, maximize ${E_{P(\mathbf{\hat{Z}}_{}^{m}, \mathbf{Z}_{}^{{m}'})} (-\log{Q(\mathbf{\hat{Z}}_{}^{m}|\mathbf{Z}_{}^{{m}'})})}$ is equivalent to minimize the mean squared error between $\mathbf{\hat{Z}}_{}^{m}$ and $\mathbf{Z}_{}^{m}$. Therefore, the reconstruction loss $L_{recover}$ can be rewritten as 
\begin{equation} \label{eq:recover_loss}
\frac{1}{N} \sum_{i=1}^{N} \left[ \left\| \hat{z}_{i}^{m} - {z}_{i}^{{m}} \right\|_{2}^{2}
+ {\lambda_r} H(p({\hat{y}_i}|{\hat{z}}_{i}^{m},{z}_{i}^{{m}'}), p({y_i}|{z}_{i}^{m},{z}_{i}^{{m}'})) \right]
\end{equation}

\subsubsection{\textbf{Meta training framework}} \label{sec:meta_training}
By training the feature mapping network to learn the optimal parameters $\mathbf{W}^{m}$, the distribution transformation could better capture the relationships between modalities, thereby achieving more accurate modality reconstruction. 
However, severe modality missing causes the data distribution mismatch with the training data, requiring the feature mapping network to extract more generalized distribution relationships applicable to {various feature distributions}.
Moreover, image distortion increases the diversity and inconsistency of features, imposing higher demands on the alignment of consistent semantic representations across different modalities.
As discussed in Sec.~\ref{sec:mtv_label}, the amount of labeled data is limited for training, hence improving the model ability to learn the distribution mapping relationships within limited training data becomes more challenging.

\newrev{As an emerging paradigm recently, meta learning~\cite{finn2017model, gui2018few, sun2019meta, chen2021meta} aims to guide the model in acquiring prior knowledge on how to learn new knowledge with only a few training samples.}
Rather than training on a single \rev{data distribution}, the model is exposed to a diverse set of data distributions, allowing it to recognize the common patterns across them. 
This exposure enables the model to leverage similar patterns when encountering new data distributions, thereby reducing the need for extensive training data.
Taking its advantage of the fast learning capacity in few-shot training~\cite{gui2018few, sun2019meta, chen2021meta}, we adopt meta-training to learn the informative mapping relationship among different modality distributions efficiently.

To learn feature mapping relationships effectively, \name evaluates the performance of modality reconstruction network on {various batches of modality-complete samples within limited labeled data}. The modality reconstruction model seeks to minimize the overall loss across those data according to $L_{recover}$ in Eqn.~\eqref{eq:recover_loss}, thereby learning generalized distribution mapping relationships that facilitate quick adaptation to new data and efficient recovery across diverse distributions.

Therefore, during meta-training, we encourage the semi-supervised learning framework to identify distinctive features of different status \rev{in the presence of modality missing}.
The semi-supervised learning framework is first initialized with the current parameters of the modality reconstruction network from a well-optimized starting point, and then updates its parameters according to $L_{ssl}$ in Eqn.~\eqref{eq:ssl_loss} using both the available data and the recovered data from the missing modality. This initialization ensures that the learned cross-modality relationships are leveraged for data recovery, providing pre-trained knowledge to enhance the generalization of status recognition from limited labeled data.
Finally, the modality reconstruction model is evaluated with the updated parameters of the semi-supervised learning model, and the overall loss for jointly optimizing both models is expressed as
\begin{equation}
    L_{all}=L_{ssl}+L_{recover}
\end{equation}

The complete training procedure of our \name is outlined in Algorithm.~\ref{alg:iv3m}.
With the limited labeled data, we apply the meta-training framework to simultaneously optimize the semi-supervised learning model and the modality reconstruction network with the overall loss $L_{all}$, exploring the discriminative information for status classification with accurate data recovery. 
For the missing data, we first utilize the modality reconstruction module to predict its feature representations, and then apply to the proposed confidence-based semi-supervised learning framework. The high-confidence pseudo labels are selected by the adaptive threshold adjustment which mitigates class imbalance, while the remaining low-confidence unlabeled data are fully utilized with contrastive learning, facilitating feature extraction for informative feature representation.

\RestyleAlgo{ruled}
\LinesNumbered
\begin{algorithm}
\caption{Our \name training.}\label{alg:iv3m}
\setstretch{0.8}
\small
\SetKwInOut{Input}{Require}
\SetKwProg{Fns}{Main}{:}{}
\SetKwFunction{Fns}{Main}
\SetKwProg{Fn}{}{:}{}
\SetKwFunction{MR}{Modality Reconstruction}
\SetKwFunction{MU}{Model Update}

\KwData{$\mathcal{D}_{l}$ and $\mathcal{D}_{u}$ are the training datasets for labeled and unlabeled data with $N_l$ and $N_u$ samples, respectively.}

{
\For{$t$ in epochs}{
    \For{$b$ in $batch\_count$}{
    \% Meta-training for classification and modality reconstruction\;
    \For{labeled data $x_i \in D_l$}{
    $L_{cls} \leftarrow$ Eqn.~\eqref{eq:classification_loss}\;
    $L_{recover} \leftarrow$ {Modality Reconstruction} ($\{z_i,y_i\}$)\;
    }
    \% Pseudo-labeling with data recovery\;
    \For{unlabeled data $x_i \in D_u$}{
    \uIf{$m$-th modality missing in $x_i$}{
    $z_i^{m'} \leftarrow$ {Feature Encoder} ($x_i^{m'}$), $\forall m'\ne m, m'\in \left\{ 1, ..., M \right\}$\; 
    $z_i^m \leftarrow$ {Feature Mapping} ($z_i^{m'}, \mathbf{V}^{m}$)\;
    }
    \uElse{
    $z^m_i \leftarrow$ {Feature Encoder} ($x^m_i$), $\forall m=1,...,M$\;
    }
    
    \uIf{$\max(p(y_i|a(z_i^m, z_i^{m'}))) \ge \tau_t(c)$}{
    $\hat{y}_i = {\rm argmax}(p(y_i|a(z_i^m, z_i^{m'})))$\;
    $L_{pl} \leftarrow$ Eqn.~\eqref{eq:consistency_loss}\;
    $\sigma_t^u(c) = \sigma_t^u(c) + (\hat{y}_i==c), \forall c=1,...,C$\;
    }
    \uElse{
    $L_{con} \leftarrow$ Eqn.~\eqref{eq:contrastive_loss}\;
    }
    }
    $L_{all} = L_{cls} + \lambda_p L_{pl} + \lambda_c L_{con} + L_{recover}$\;
    }
    $\tau_{t+1}(c) = \min\{(p_t(c) + \tau - D_{KL}(p_t)), \tau_h\}$
}
}

\end{algorithm}

\vspace{-0.2cm}
\section{Implementation and Experimental Setup} 
\label{sec:implementation}
In this section, we demonstrate the detailed implementation of our \name system in the in-car status monitoring application using {Stressors} dataset~\cite{taamneh2017multimodal} and Drive\&Act dataset~\cite{martin2019drive}. The performance of \name is evaluated against several multimodal monitoring algorithms with specific hyper-parameter settings.

\paragraph{Dataset and tasks}
We use two public datasets in our experiment, the {Stressors} dataset~\cite{taamneh2017multimodal} for stress detection and the Drive\&Act dataset~\cite{martin2019drive} for activity recognition.
The {Stressors} dataset~\cite{taamneh2017multimodal} is a multimodal dataset designed to analyze drivers' status under various cognitive, emotional, and sensorimotor stressors, where the stress-related conditions allow for identifying whether subjects feel stressful in the car. {\needrev{Since experiencing stress often exhibits patterns similar to the early signs of abnormal status as stated in Sec.~\ref{sec:sd_overview}}, we utilize the facial information, physiological signals, and vehicle data to detect stress status for the driver and recognize abnormal health conditions for passengers using facial and physiological data.}
To continuously record status data of subjects, 
facial information is captured at a frame rate of 25 fps while the physiological signals (such as heart rate and breathing rate) and the vehicle-related parameters (such as speed, acceleration, brake force, steering angle, and lane position) are measured at a sampling rate of 1Hz.
\rev{Here we shift the facial images with different angles to simulate the camera rotation.}
The time sequences of the multimodal data are synchronized using the global timestamp and segmented into {120-second windows with a 60-second overlap as input data.}
Due to the absence of ground truths, labels are manually created based on facial video and palm EDA sensor data. 
It is worth to note that there are considerable amount of data lost in the original dataset, therefore, we take out a total of 300 samples from 24 subjects, using 1 subject's data to create the test dataset with 30 samples and 90\% and 10\% of other 23 subjects' data for training and validation, respectively.

{Activities often involve dynamic behaviors that require an understanding of temporal sequences and spatial movements, making them more challenging than status recognition. Therefore, we further evaluate the performance of our \name on activity recognition.
The Drive\&Act dataset~\cite{martin2019drive} is a multimodal driving behavior dataset to recognize distractive activities in autonomous vehicles. The dataset provides detailed annotations for 34 fine-grained driver behaviors, such as fetching object, looking around, and talking on phone. Here we use RGB images from the right-top view and 3D skeleton in our experiments, where RGB video is sampled at 30 fps and 3D pose at 25 Hz. We segment data streams into individual time-space frames using a 120-second recording window with a 60-second overlap. 75\% and 15\% of 15 subjects' data are used for training and testing, respectively.}

\paragraph{Baselines}
To investigate the advantages of our \name framework, we compare it with other benchmarks:
\begin{itemize}
    \item {\bf{DeepSense~\cite{yao2017deepsense}}} is a popular supervised learning network for general multimodal sensing applications like driver monitoring, which integrates convolutional and recurrent neural networks to extract robust and distinct features according to interactions among similar mobile sensors. 
    \item {\bf{Contrastive Multi-view Learning (CMC)~\cite{tian2020contrastive}}} is a state-of-the-art contrastive learning framework for multi-view multimodal predictive learning, which extracts the consistent feature among multiple sensors with pretrained feature encoders from unlabeled multimodal data.
    \item {\bf{Missing Modality Imagination Network (MMIN)~\cite{zhao2021missing}}} is a unified model for multimodal emotion recognition with uncertain missing modality, where two independent networks are employed to reconstruct the missing modality based on other available modalities in the forward direction and also predict the available modalities based on the imagined missing modality in the backward direction.
    \item {\bf{FlexMatch~\cite{zhang2021flexmatch}}} is a cutting-edge semi-supervised learning framework which incorporates an adaptive strategy for label generation. The threshold for pseudo-labeling is adjusted based on learning status of each class, facilitating feature extraction from classes with learning difficulties. 
\end{itemize}

\paragraph{Hardware and hyper-parameters}
We train our \name on a server with an NVIDIA Quadro RTX 3000 GPU, Intel i9-10885H CPUs, and 256 GB RAM. Python 3.7 and PyTorch 1.9.1 are used for implementing the application.
{The facial information, physiological signals, and vehicle data are processed to dimensions of (60, 50, 8), (120, 8), and (120, 6) for training, respectively, with the extracted feature dimension of 1280.}
We employ the same LSTM based feature encoders \rev{with a modality-complete base model} for MMIN as described in~\cite{zhao2021missing}. For our \name and other baselines, we use a CNN network with 6 layers for feature extraction, 4 linear layers for classification, and 1 linear layer for modality reconstruction.
We set the optimizer, learning rate and batch size as Adam, 0.0001 and 8, respectively. The number of important principal components, the number of negative samples, and the weight parameters of overall training loss are set to $K=4$, $S=8$, $\lambda_c=0.1$, and $\lambda_p=0.1$, respectively. 
{We assume that the labeled data constitutes 10\% of the total training samples, with all modalities available, and the unlabeled dataset contains 90\% incomplete samples 
with facial data missing by default.}
We run each experiment five times and report the average performance on the testing set.


\section{Evaluation} \label{sec:evaluation}
In this section, we conduct the overall performance evaluation of our \name and various benchmarks in comparison of detection accuracy, precision, recall and convergence time. Additionally, we evaluate the performance of the proposed framework under different labeling rates and data missing rates, 
The contributions of different modules within the our \name framework are also analyzed to illustrate their individual roles in the overall design.

\vspace{-0.2cm}
\subsection{The Overall Performance of our \name}

\begin{figure}[t]
\centering
\subfloat[Stressors \label{fig:accuracy_label_stressor}]{
    \includegraphics[width=0.495\linewidth]{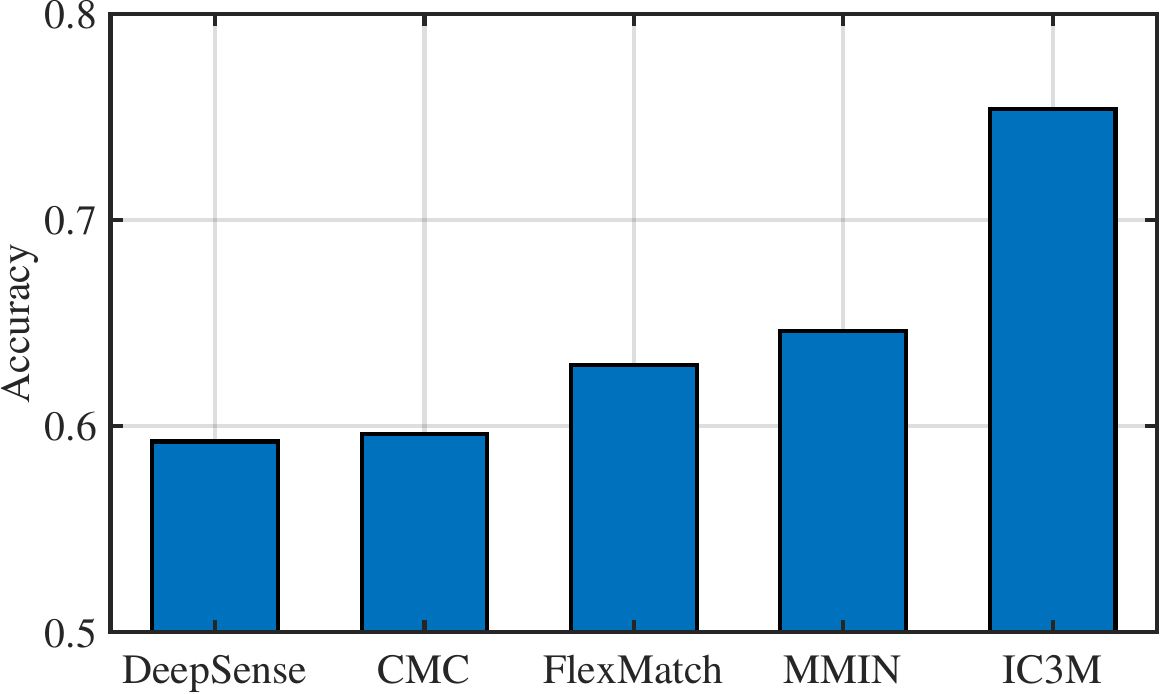}
}
\subfloat[Drive\&Act \label{fig:accuracy_label_UTD}]{
    \includegraphics[width=0.495\linewidth]{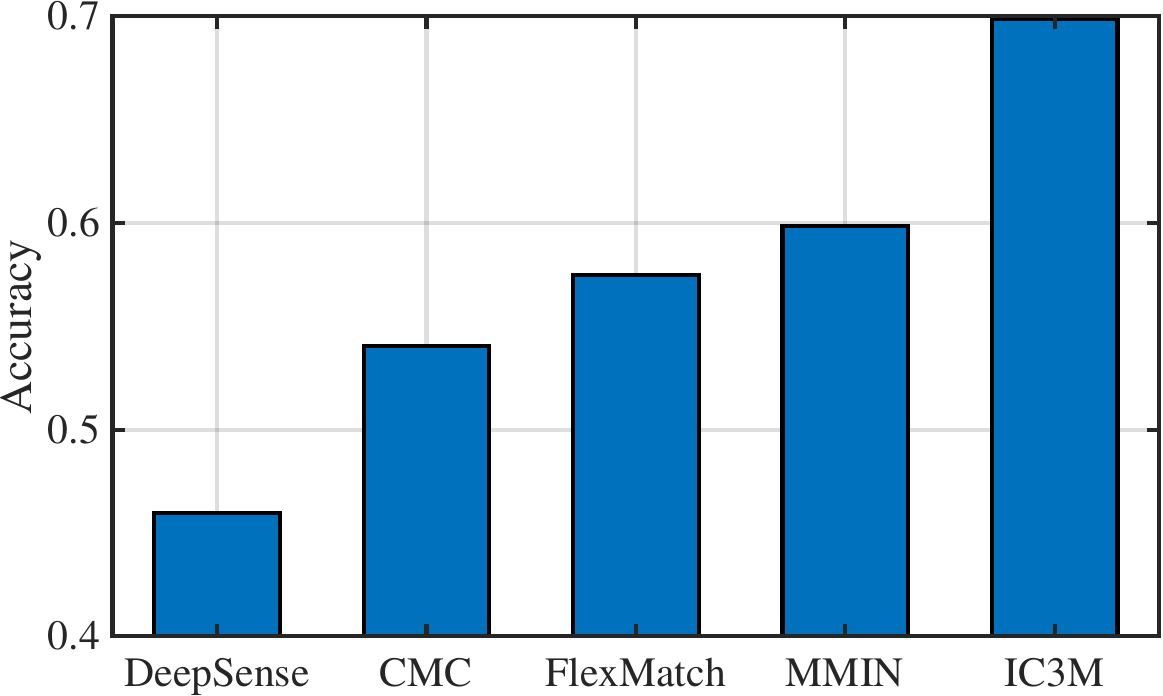}
}
\caption{Accuracy for baselines with 90\% facial information lost and 10\% labelled data.}
\label{fig:accuracy_label}
\vspace{-2ex}
\end{figure}

\subsubsection{Detection Accuracy}
Fig.~\ref{fig:accuracy_label} shows the status detection accuracy for our \name and other benchmarks on the {Stressors} and Drive\&Act datasets with 90\% facial data missing and 10\% labeled data.
It is clear to see that the proposed \name framework outperforms all other baselines. {This superiority mainly stems from the better utilization of pseudo labels and well-recovered data, which enhances the feature extraction for status classification}.
By selecting the high-confidence pseudo labels for training and further leverage the low-confidence pseudo labels for effective feature extraction, the proposed \name outpaces the accuracy of MMIN and DeepSense by \needrev{11\% and 16\%} for Stressors dataset and \needrev{10\% and 23\%} for Drive\&Act dataset, respectively.
Although FlexMatch and CMC could also leverage the large number of unlabeled data for performance enhancement, their detection accuracy is below \needrev{64\%} as the lack of data imputation strategy limits the valuable information utilized in model training, making model struggle to capture informative features.
In contrast, our \name achieves accurate data recovery by restoring the complementary information from missing modality, thereby providing more information to explore discriminative features for reliable status classification.

\begin{table}[t]
\centering
\scalebox{0.8}{
\begin{tabular}{ |c||c||c|c|c|c|c| }
    \hline
    \multirow{2}{*}{Dataset} &\multirow{2}{*}{Metrics} &\multicolumn{5}{c|}{Model Performance} \\ 
    & &DeepSense &CMC &FlexMatch &MMIN &\name \\
    \hline
    \multirow{4}{*}{Stressors} 
    &\textit{Accuracy}  & 59.25  & 59.63  & 63.98  & 64.62  & 75.37 \\
    &\textit{Precision} & 66.70  & 68.59  & 62.71  & 52.46  & 89.13 \\
    &\textit{Recall}    & 84.13  & 69.51  & 60.48  & 49.72  & 76.78 \\
    &\textit{F1-score}  & 39.65  & 54.15  & 59.57  & 40.82  & 71.15 \\
    \hline
    \multirow{4}{*}{Drive\&Act} 
    &\textit{Accuracy}  & 45.96  & 54.07  & 57.51  & 59.78  & 69.89 \\
    &\textit{Precision} & 41.57  & 61.79  & 59.59  & 51.21  & 77.98 \\
    &\textit{Recall}    & 68.05  & 66.32  & 57.02  & 45.64  & 72.03 \\
    &\textit{F1-score}  & 30.09  & 48.91  & 53.19  & 38.33  & 66.24 \\
    \hline
\end{tabular}
}
\caption{Class-specific metrics for baselines with 90\% facial information lost and 10\% labelled data.}
\label{tab:accuracy_metrics}
\vspace{-2ex}
\end{table}

\subsubsection{Class-specific Metrics}
{Since the unreasonable high accuracy can be achieved by ignoring the minority class in imbalanced dataset, we also compare the precision, recall and f1 score for status identification}, which are class-specific metrics to evaluate the recognition performance on both normal status and abnormal status.
As shown in Table~\ref{tab:accuracy_metrics}, our \name achieves precision above \needrev{78\%}, indicating its ability to accurately recover missing data and reliably classify status even with missing modalities. 
With the help of balanced data distribution and low-confidence pseudo labels, the proposed adaptive threshold pseudo-labeling focuses on capturing informative features from sparse abnormal status class, thus our \name outperforms CMC, FlexMatch and MMIN in recall by about \needrev{6\%, 15\% and 26\%}, respectively.
We also notice that our \name has a significantly higher F1 score over \needrev{66\%}, demonstrating its superior performance in sensitivity and reliability for timely identifying in-car abnormal status.
Moreover, it is noteworthy to observe that although the detection accuracy is over 60\% for FlexMatch and MMIN, the precision and recall are notably lower than our \name. 
{Due to the overlook of class imbalance, which limits the feature representation learning for abnormal status, FlexMatch and MMIN struggle to identify samples in abnormal status, thus resulting in inferior precision and recall for abnormal status detection.}

\subsubsection{Convergence Performance}
\begin{figure}[t]
\centering
\subfloat[Stressors \label{fig:accuracy_converge_stressor}]{
    \includegraphics[width=0.495\linewidth]{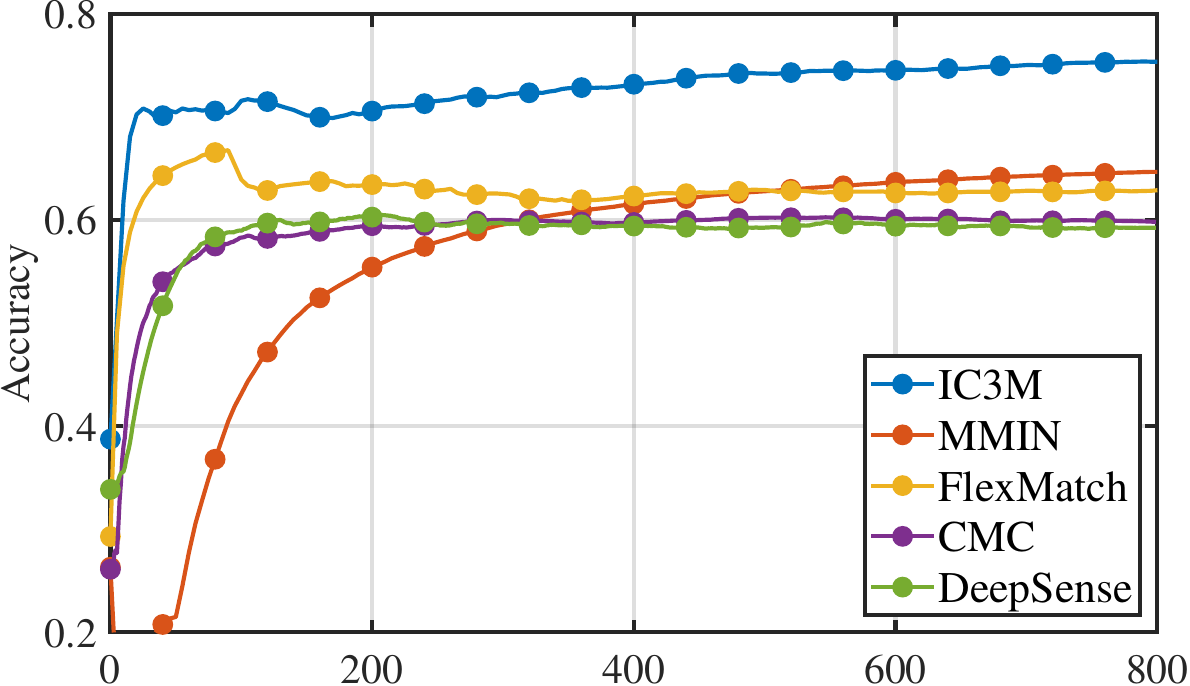}
}
\subfloat[Drive\&Act \label{fig:accuracy_converge_UTD}]{
    \includegraphics[width=0.495\linewidth]{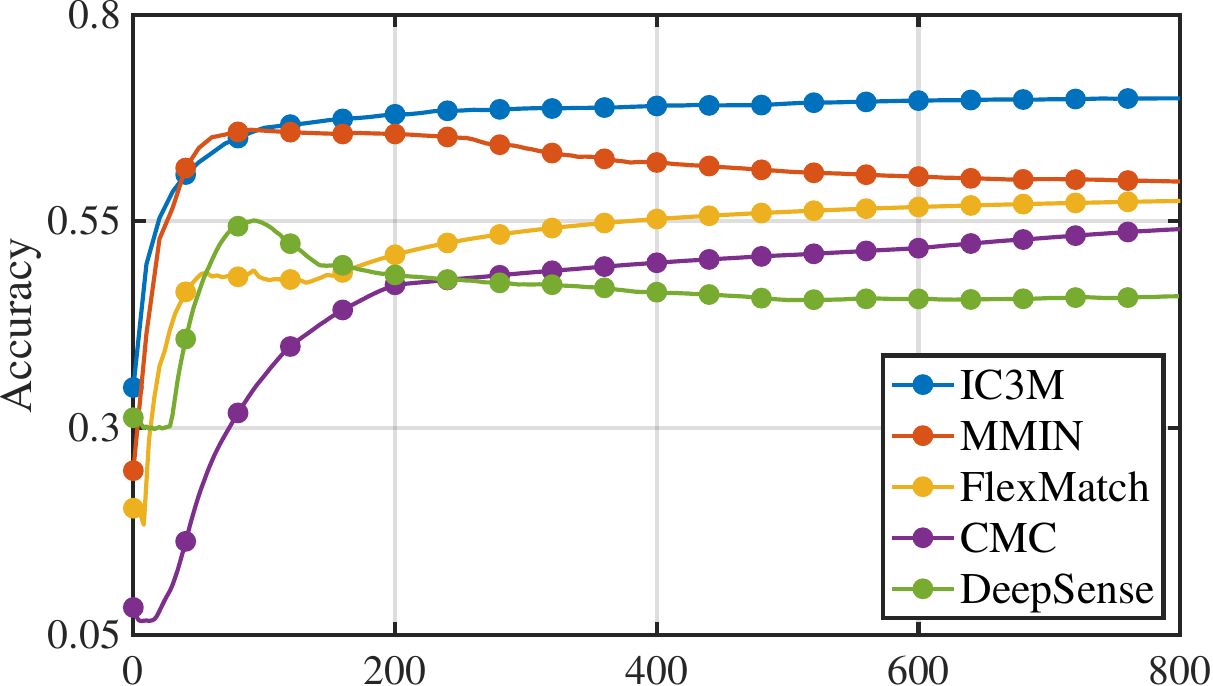}
}
\caption{Convergence for baselines with 90\% facial information lost and 10\% labelled data.}
\label{fig:accuracy_convergence}
\vspace{-2ex}
\end{figure}
Fig.~\ref{fig:accuracy_convergence} compares the convergence of our \name with four baselines on the Stressors and Drive\&Act datasets when training with 90\% facial information lost and 10\% labelled data.
It is noticed that our \name converges much faster and requires only \needrev{100} epochs to achieve the highest testing accuracy, while MMIN and DeepSense take \needrev{400 and 200} epochs, respectively, as they struggle to capture informative features from the limited training samples with modality missing under the supervised learning approach.
Taking advantage of large amounts of unlabeled data for feature extraction, CMC achieves a comparable convergence rate to \name on Stressors dataset. However, the imbalanced class distribution and modality missing limits its detection performance.
Moreover, our \name demonstrates even faster convergence speed than FlexMatch, primarily attributing to the adaptive threshold adjustment that keeps the class balance in training and the leverage of low-confidence data that enhances feature extraction via contrastive learning. 
FlexMatch, on the other hand, neglects class imbalance when adjusting threshold, thus hindering the feature learning for abnormal status class and affecting training efficiency.

\vspace{-0.2cm}
\subsection{Optimal Hyper-parameters}

\begin{figure}[t]
\centering
\subfloat[Accuracy \label{fig:label_rate_acc}]{
    \includegraphics[width=0.495\linewidth]{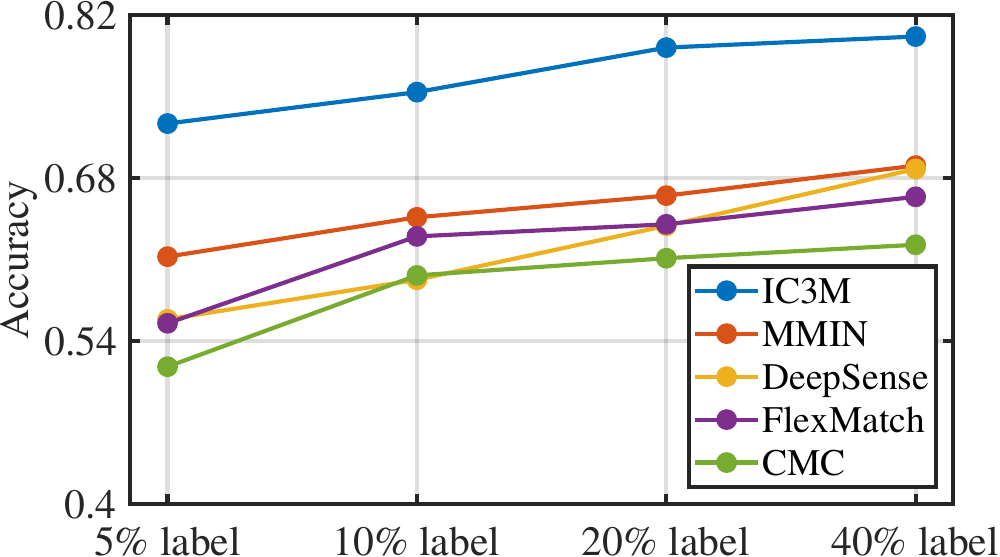}
}
\subfloat[Convergence \label{fig:label_rate_time}]{
    \includegraphics[width=0.49\linewidth]{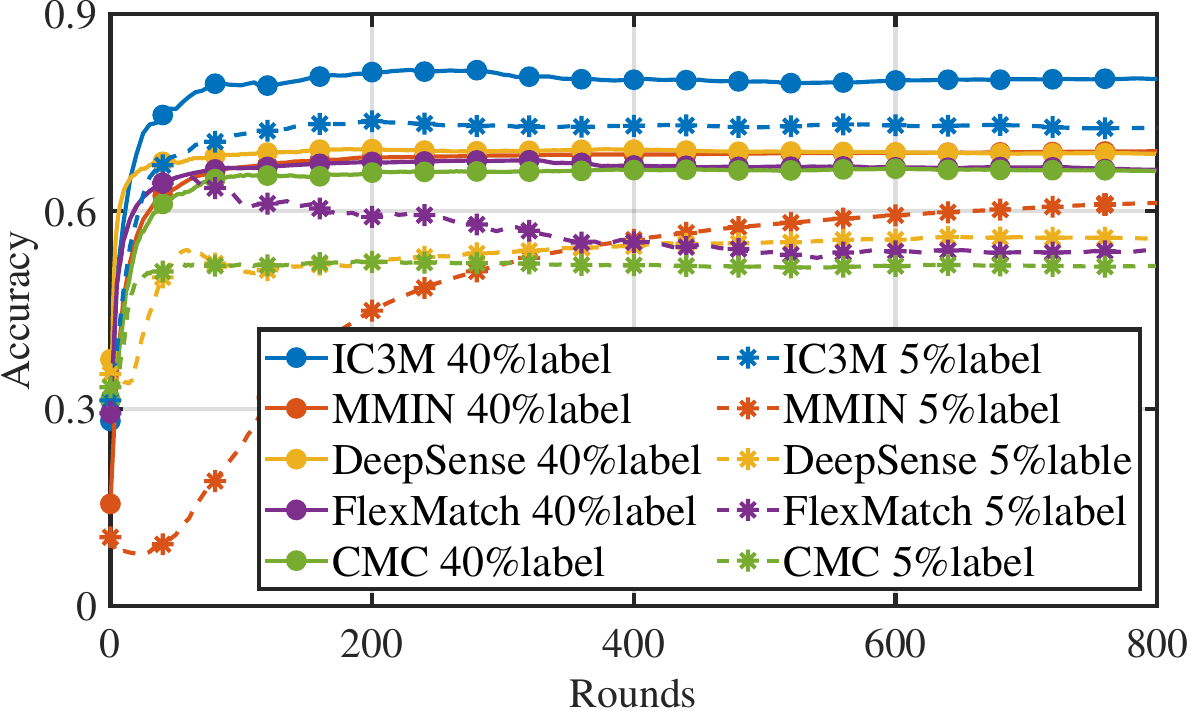}
}
\caption{Model performance with different proportion of labeled data under 90\% facial information missing.}
\label{fig:label_rate}
\vspace{-2ex}
\end{figure}

\begin{table}[t]
\centering
\scalebox{0.8}{
\begin{tabular}{ |c||c||c|c|c|c|c| }
    \hline
    \multirow{2}{1.1cm}{Labeling} &\multirow{2}{*}{Metrics} &\multicolumn{5}{c|}{Model Performance} \\ 
    & &DeepSense &CMC &FlexMatch &MMIN &\name \\
    \hline 
    \multirow{4}{1.1cm}{5\% labels} 
    &\textit{Accuracy}  & 55.86  & 51.79  & 55.52  & 61.24  & 72.67 \\
    &\textit{Precision} & 72.40  & 79.84  & 53.75  & 50.65  & 80.68 \\
    &\textit{Recall}    & 65.13  & 41.05  & 62.04  & 51.45  & 79.92 \\
    &\textit{F1-score}  & 47.36  & 51.50  & 54.62  & 40.74  & 67.18 \\
    \hline
    \multirow{4}{1.1cm}{10\% labels} 
    &\textit{Accuracy}  & 59.25  & 59.63  & 62.98  & 64.62  & 75.37 \\
    &\textit{Precision} & 66.70  & 73.59  & 62.71  & 52.46  & 89.13 \\
    &\textit{Recall}    & 84.13  & 69.51  & 60.48  & 49.72  & 76.78 \\
    &\textit{F1-score}  & 39.65  & 54.15  & 59.57  & 40.82  & 71.15 \\
    \hline
    \multirow{4}{1.1cm}{20\% labels} 
    &\textit{Accuracy}  & 63.89  & 62.26  & 64.02  & 66.48  & 79.19 \\
    &\textit{Precision} & 72.40  & 71.02  & 64.27  & 53.12  & 97.35 \\
    &\textit{Recall}    & 89.59  & 76.28  & 62.04  & 50.78  & 72.25 \\
    &\textit{F1-score}  & 39.16  & 56.36  & 60.46  & 41.04  & 77.82 \\
    \hline
    \multirow{4}{1.1cm}{40\% labels} 
    &\textit{Accuracy}  & 68.77  & 66.11  & 66.38  & 69.06  & 80.14 \\
    &\textit{Precision} & 70.99  & 76.09  & 69.26  & 53.98  & 97.44 \\
    &\textit{Recall}    & 93.58  & 67.52  & 65.91  & 49.59  & 73.20 \\
    &\textit{F1-score}  & 48.59  & 57.06  & 61.49  & 45.03  & 79.04 \\
    \hline
\end{tabular}
}
\caption{Performance comparison for baselines with different labeling rate under 90\% facial information missing.}
\label{tab:label_rate}
\vspace{-2ex}
\end{table}

\subsubsection{The Impact of Labeling Rate}
Fig.~\ref{fig:label_rate} and Table~\ref{tab:label_rate} present the performance of status detection for our \name and other benchmarks with different proportion of labeled data on the Stressors dataset under 90\% facial information missing.
We observe that our \name consistently exhibits the best detection performance, particularly when trained with a very small portion of labeled samples. 
Due to accurate modality reconstruction with the customized design for pseudo-labeling and unlabeled data utilization, our \name achieves a detection accuracy of \needrev{72.67\%} with \needrev{80.68\% precision and 79.92\% recall} using just 5\% labels. 
Conversely, the absence of balanced class distribution and sufficient recovered samples in other benchmarks hinders the effectiveness of feature extraction, resulting in accuracy below \needrev{62\%} and slower convergence with 5\% labeled data.
Moreover, the comprehensive comparison of various evaluation metrics in Table~\ref{tab:label_rate} illustrates that the proposed \name achieves comparable performance of other baselines with 40\% data labelled, when trained with only 5\% labels.
Compared with other baselines, our design of adaptive threshold training balances the label quantity for model training with further explored information from unlabeled data, thereby facilitating the informative feature extraction when training with limited labeled data. 


\begin{figure}[t]
\centering
\subfloat[Accuracy \label{fig:missing_rate_acc}]{
    \includegraphics[width=0.495\linewidth]{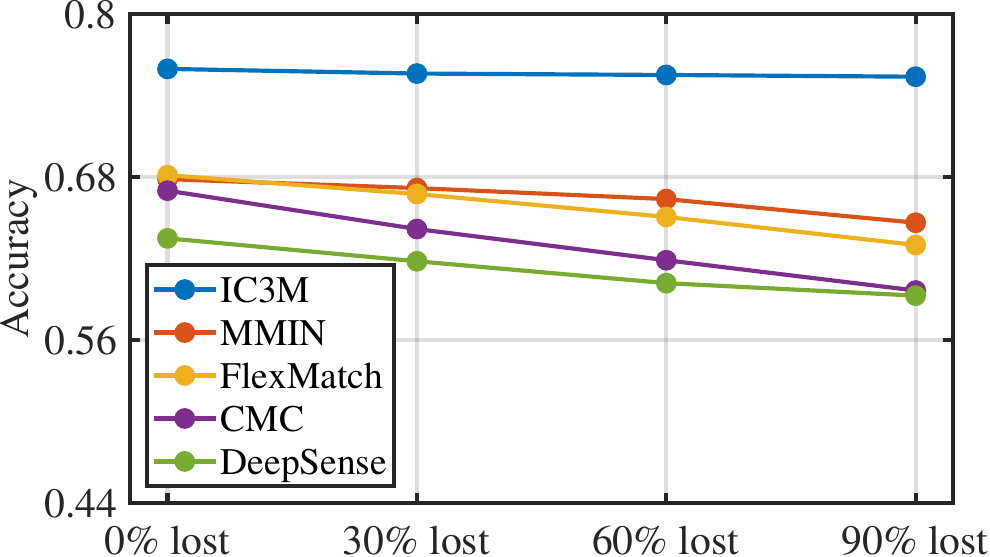}
}
\subfloat[Convergence \label{fig:missing_rate_time}]{
    \includegraphics[width=0.49\linewidth]{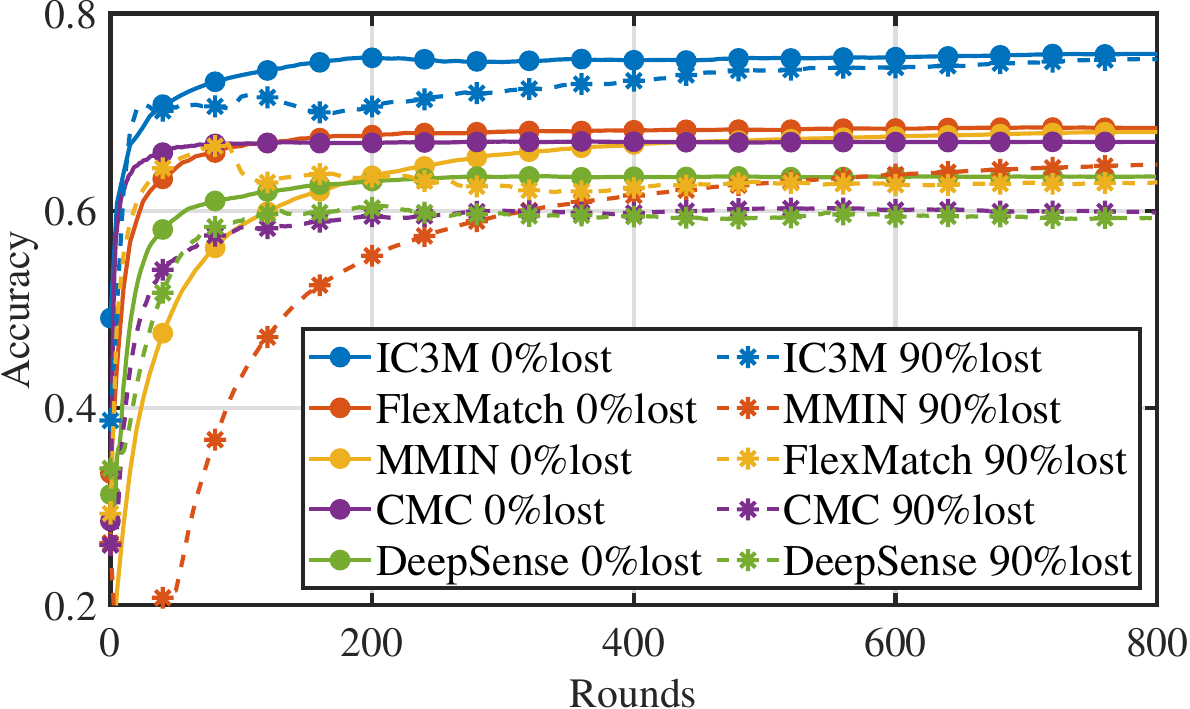}
}
\caption{Model performance with different proportion of facial information missing under 10\% labeling rate.}
\label{fig:missing_rate}
\vspace{-2ex}
\end{figure}

\begin{table}[t]
\centering
\scalebox{0.8}{
\begin{tabular}{ |c||c||c|c|c|c|c| }
    \hline
    \multirow{2}{1.0cm}{Missing} &\multirow{2}{*}{Metrics} &\multicolumn{5}{c|}{Model Performance} \\ 
    & &DeepSense &CMC &FlexMatch &MMIN &\name \\
    \hline 
    \multirow{4}{1.0cm}{0\% lost} 
    &\textit{Accuracy}  & 63.47  & 66.98  & 68.12  & 67.83  & 75.94 \\
    &\textit{Precision} & 67.07  & 90.89  & 65.55  & 57.86  & 89.53 \\
    &\textit{Recall}    & 86.59  & 69.68  & 68.26  & 51.34  & 77.72 \\
    &\textit{F1-score}  & 42.23  & 59.08  & 64.78  & 41.86  & 72.35 \\
    \hline
    \multirow{4}{1.0cm}{30\% lost} 
    &\textit{Accuracy}  & 61.78  & 64.15  & 66.73  & 67.17  & 75.60 \\
    &\textit{Precision} & 65.16  & 88.14  & 62.04  & 57.90  & 89.19 \\
    &\textit{Recall}    & 85.50  & 69.87  & 68.48  & 50.17  & 77.28 \\
    &\textit{F1-score}  & 39.66  & 58.22  & 62.15  & 41.42  & 72.06 \\
    \hline
    \multirow{4}{1.0cm}{60\% lost} 
    &\textit{Accuracy}  & 60.17  & 61.86  & 65.05  & 66.36  & 75.46 \\
    &\textit{Precision} & 66.49  & 82.94  & 64.56  & 54.48  & 89.07 \\
    &\textit{Recall}    & 83.30  & 68.77  & 64.73  & 49.63  & 77.07 \\
    &\textit{F1-score}  & 39.70  & 55.28  & 60.91  & 41.11  & 71.91 \\
    \hline
    \multirow{4}{1.0cm}{90\% lost} 
    &\textit{Accuracy}  & 59.25  & 59.63  & 62.98  & 64.62  & 75.37 \\
    &\textit{Precision} & 66.70  & 73.59  & 62.71  & 52.46  & 89.13 \\
    &\textit{Recall}    & 84.13  & 69.51  & 60.48  & 49.72  & 76.78 \\
    &\textit{F1-score}  & 39.65  & 54.15  & 59.57  & 40.82  & 71.15 \\
    \hline
\end{tabular}
}
\caption{Performance comparison for baselines with different missing rate under 10\% labeled data.}
\label{tab:missing_rate}
\vspace{-2ex}
\end{table}

\subsubsection{The Impact of Modality Missing Rate}
Since the facial information is frequently missing in the {Stressors} dataset, we investigate the performance of status detection with different proportion of facial data missing on the Stressors dataset under 10\% labeling rate in Fig.~\ref{fig:missing_rate} and Table~\ref{tab:missing_rate}. 
It is illustrated in Fig.~\ref{fig:missing_rate} that our \name achieves the highest detection accuracy in different modality missing rates with the stablest performance compared to other benchmarks.
This is due to {the complementary information provided by the recovered modality, which enhances status classification and thereby improving recognition performance.}
The fast learning capacity of meta learning further enhances the model generalization to different data distributions, allowing our \name to maintain over \needrev{75\%} accuracy with only a slight drop of \needrev{0.6\%}, with 90\% modality data missing compared to no missing data (see Table~\ref{tab:missing_rate}).
In contrast, DeepSense, CMC, and FlexMatch struggle to extract important features without a data recovery method, resulting in a drastic drop in accuracy by over \needrev{4\%} and F1-score by nearly \needrev{5\%} when 90\% of the modality data is missing.
Although MMIN can also recover missing data from other modalities, its effectiveness is limited by the negligence of modality distribution gap, further highlighting the superior adaptability of our \name under severe modality data missing and camera rotation. 

\vspace{-0.2cm}
\subsection{Ablation Study}

\subsubsection{{Adaptive Threshold Pseudo-Labeling}}
\begin{figure}[t]
\centering
\subfloat[Accuracy \label{fig:adaptive_ssl_acc}]{
    \includegraphics[width=0.495\linewidth]{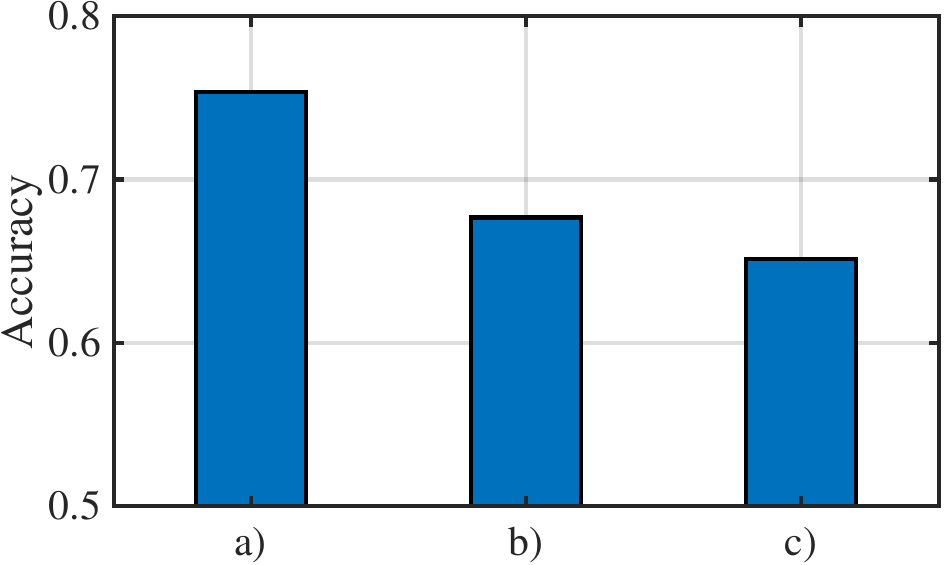}
}
\subfloat[Convergence \label{fig:adaptive_ssl_time}]{
    \includegraphics[width=0.49\linewidth]{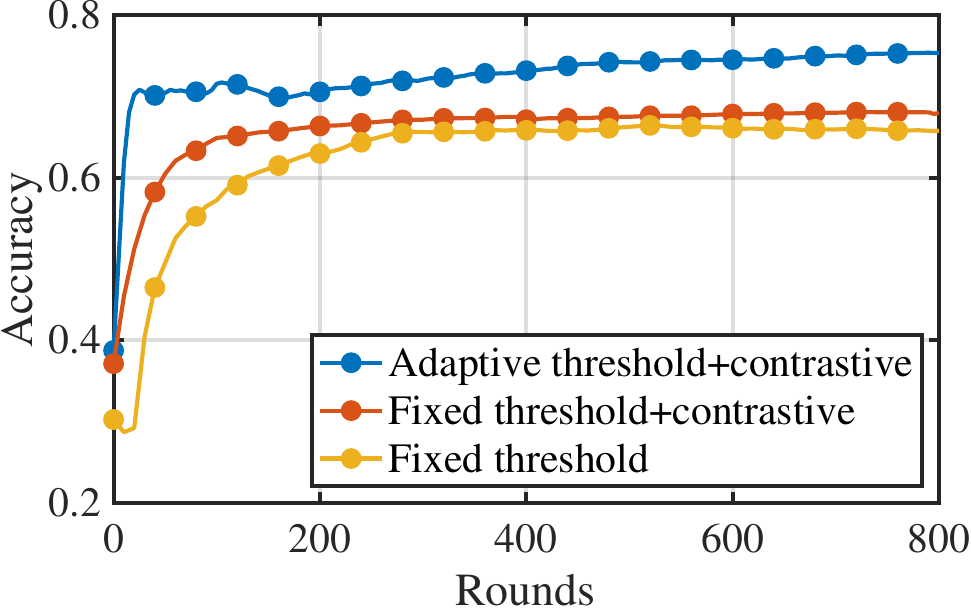}
}
\caption{Model performance of adaptive threshold training with 90\% facial information missing and 10\% labeling rate}.
\label{fig:adaptive_ssl}
\vspace{-2ex}
\end{figure}

Fig.~\ref{fig:adaptive_ssl} presents the impact of adaptive threshold pseudo-labeling on the Stressors dataset with 90\% facial information missing and 10\% labeled data. 
{We evaluate the performance of adaptive threshold pseudo-labeling by comparing model training under three conditions:} a). adaptive threshold adjustments with the proposed contrastive loss, b). the fixed threshold with the proposed contrastive loss, and c). the fixed threshold without contrastive learning.
The results show that the proposed adaptive confidence training framework improves model performance and training efficiency, achieving \needrev{70\%} accuracy in just \needrev{30} epochs.
This effectiveness is due to the selected high-confidence pseudo labels that expand the training dataset and mitigate class imbalance, enhancing feature extraction from the class with fewer samples, thus improving performance with a better learning of discriminative features from each class.
Introducing other low-confidence unlabeled data into model training through contrastive learning brings further benefits because it provides extra useful information for feature extraction from limited labels. As a result, our \name is efficient to learn informative features for abnormal status recognition with the proposed adaptive threshold pseudo-labeling framework. 



\subsubsection{Modality Reconstruction}
\begin{figure}[t]
\vspace{-.5ex}
\centering
\subfloat[Facial information missing.]{
    \includegraphics[width=0.495\linewidth]{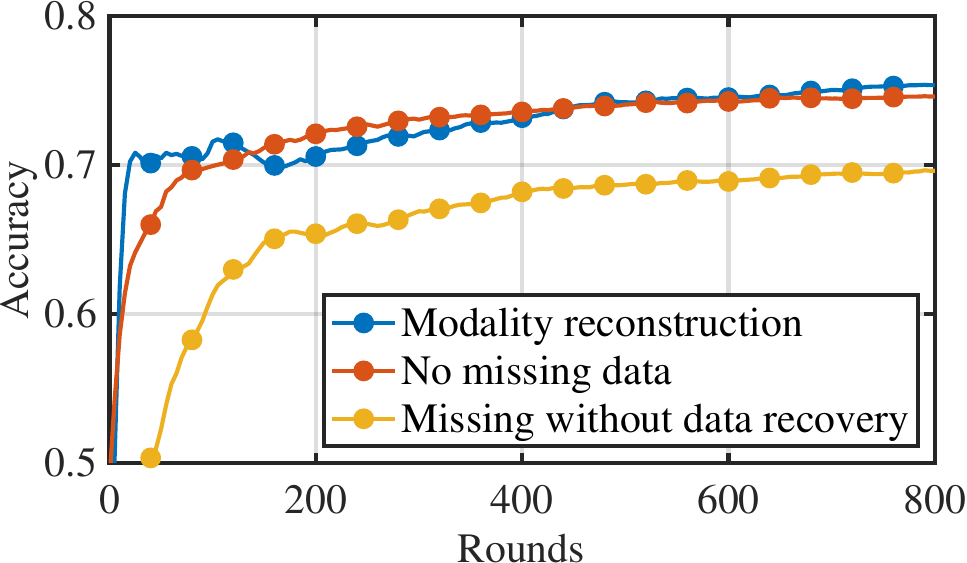}
    \label{fig:robustness_modality_facial}
}
\subfloat[Physiological signal missing.]{
    \includegraphics[width=0.495\linewidth]{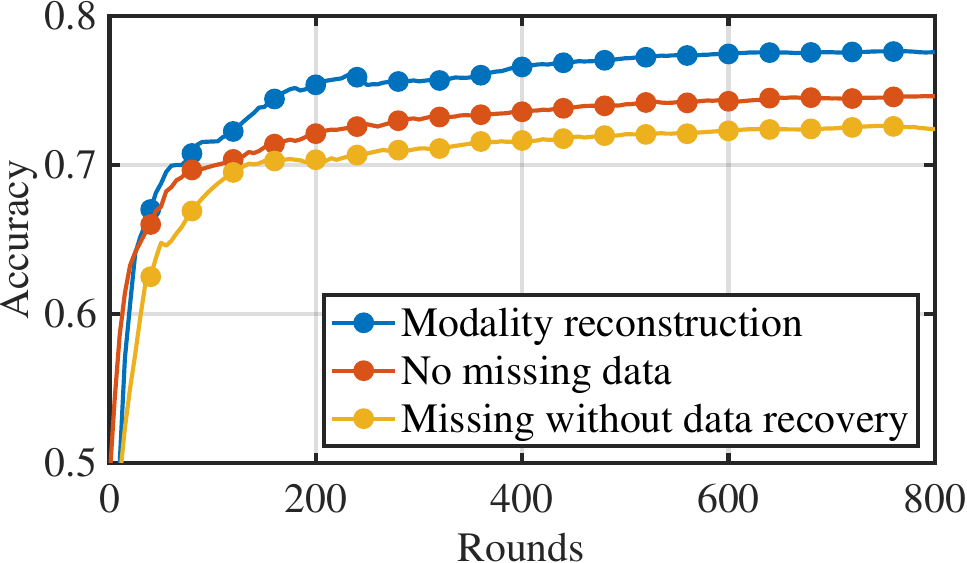}
    \label{fig:robustness_modality_vital}
}  \\
\subfloat[Vehicle data missing.]{
    \includegraphics[width=0.495\linewidth]{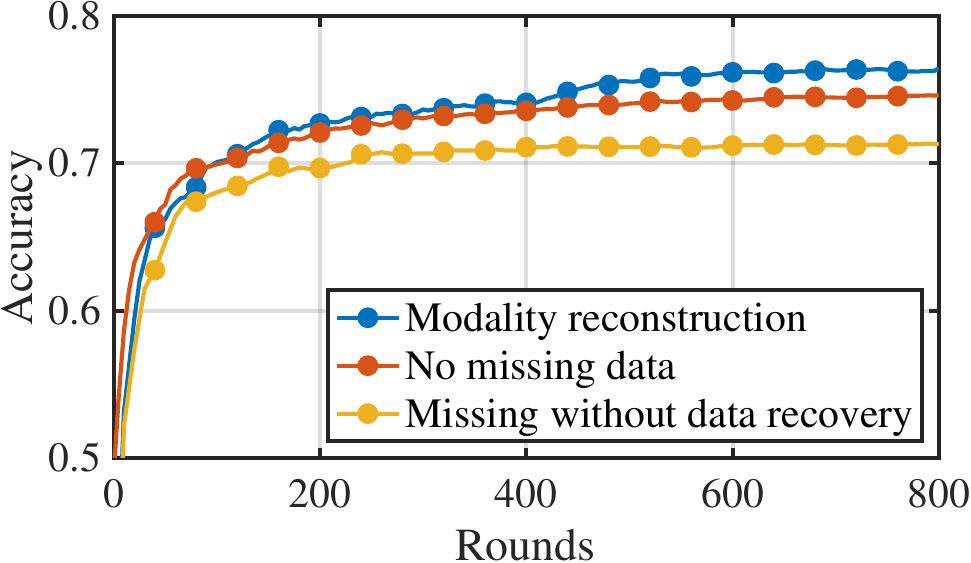}
    \label{fig:robustness_modality_vehicle}
}
\subfloat[Accuracy comparison.]{
    \includegraphics[width=0.495\linewidth]{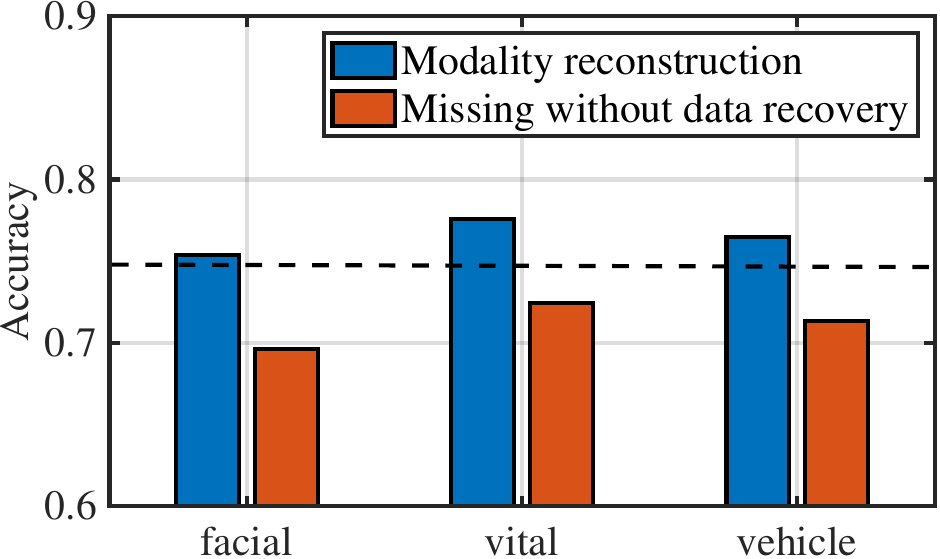}
    \label{fig:robustness_modality_statistics}
}
\caption{Model performance under different modalities missing with 10\% labeling rate and 90\% samples missing.}
\label{fig:robustness_modality}
\vspace{-2ex}
\end{figure}

Fig.~\ref{fig:robustness_modality} illustrates the impact of modality reconstruction on the Stressors dataset with 10\% labeled data and 90\% missing samples in different modality (camera, wearable sensors or vehicle data).
Our proposed modality reconstruction module outperforms the model without it, achieving nearly a \needrev{6\%} improvement in accuracy. This is owing to the benefits of the cross-modality distribution transfer, which recovers modality-specific features and provides complementary information for reliable detection.
Moreover, we observe that the difference of the performance with modality missing under different settings is \needrev{negligible}, showcasing the robustness of our approach.
It is also worth noting that with the missing modality reconstruction, our \name reaches equivalent or even better performance than without modality missing, indicating the superiority of our \name for effective multi-object monitoring under camera rotation.
The reason is that we utilize meta learning framework to learn robust features applicable to various \rev{data distributions}, improving model ability to handle unseen data and promoting its adaptability for status recognition.


\vspace{-0.2cm}
\section{Related Work} \label{sec:related_work}

{\bf{In-car multimodal monitoring: }}
Current in-car monitoring systems are designed to enhance driving safety with the integration of various sensors, such as cameras, wearable devices, and vehicle sensors (i.e., accelerator, gyroscope, and pedal), to monitor driver behaviors from multiple perspectives.
For instance, camera-based~\cite{zhang2016traffic, jegham2019mdad, rastgoo2018critical} and vehicle-based~\cite{li2021novel} driver monitoring systems capture visual information to identify dangerous behaviors such as distracted or drowsy driving, while signal-based systems~\cite{juncen2023mmdrive} utilize wearable devices to provide real-time insights into a driver’s physiological signals, such as heart rate and respiration. 
Numerous real-world experiments~\cite{yang2023aide, martin2019drive, ortega2020dmd} have also provided valuable datasets for differentiating driver status through the fusion of multiple sensing modalities.
However, most current studies in this area only focus on the status recognition of driver, largely neglecting the monitoring of passengers, which is equally critical for ensuring their safety and health. 
Although some research efforts aim to improve passenger safety, they typically focus on monitoring occupant postures to adjust the vehicle settings~\cite{abedi2022use, breed2009dynamic} or detecting whether the children are left behind accidentally~\cite{backar2022vehicle, ma2020carosense}.
Unfortunately, these studies fail to capture the fine-grained features (e.g., facial information and physiological signals), which are essential for recognizing the abnormal status of passengers.

{\bf{Training with limited labels: }}
Previous works on in-car monitoring primarily rely on supervised learning with extensive labeled data. However, in real-world in-car monitoring, obtaining such a large, well-labeled multimodal dataset is often impractical.
To address this issue, semi-supervised learning~\cite{berthelot2019mixmatch, sohn2020fixmatch, xie2020unsupervised} has emerged as a promising approach, leveraging both limited labeled data and abundant unlabeled samples to enhance model performance.
MixMatch~\cite{berthelot2019mixmatch} is a widely used SSL method that mixes labeled and unlabeled data with various augmented versions to guess low-entropy labels.
FixMatch~\cite{sohn2020fixmatch} simplifies SSL algorithms by combining consistency regularization with confidence-based pseudo-labeling through weak and strong data augmentations. 
However, these approaches often overlook the class imbalance in training datasets, where abnormal status samples are significantly fewer than normal ones. 
This imbalance makes model prone to classify data to normal status, severely impairing their ability to distinguish abnormal status. 
Recently, FlexMatch~\cite{zhang2021flexmatch} and some variants~\cite{guo2022class, wang2022freematch} attempt to mitigate this issue by {dynamically adjusting the confidence threshold for pseudo-labeling based on the model performance across different classes.}
Nevertheless, previous works always assume that the paired and complete multimodal data are available for model training and fail to incorporate the low-confidence unlabeled data in training as well.


{\bf{Modality reconstruction: }}
Handling missing modalities is a prevalent challenge in multimodal sensing systems. To address this issue, various data imputation methods have been developed, either by learning joint multimodal representations~\cite{wang2020transmodality, wang2020multimodal} or by generating missing data from other available modalities~\cite{zhao2021missing, zhou2017anomaly, zhou2021memorizing, perera2019ocgan}.
Joint multimodal representations learning aims to capture relevant information across modalities, handling incomplete modalities by \rev{knowledge transfer}.
Wang et al.~\cite{wang2020multimodal} propose a knowledge distillation-based method which trains a teacher model with modality missing and then transfers this knowledge to a student model that has access only to the incomplete modalities.
TransModality~\cite{wang2020transmodality} introduces a transformer-based approach for multimodal data recovery by aligning features across modalities through inter-modality relationships.
Generative methods, such as AutoEncoders and Variational AutoEncoders, have also been developed to predict missing modalities based on the available ones. These models extract shared information across modalities using encoders and recover data through decoders~\cite{zhao2021missing, zhou2017anomaly, zhou2021memorizing, perera2019ocgan}. 
However, these models typically focus on restoring the missing data by identifying common features across modalities, overlooking the inconsistency in feature distributions among different modalities. 
This distribution gap hinders effective feature alignment and introduces imputation noise, potentially leading to unstable results or reduced accuracy.

\vspace{-0.2cm}
\section{Conclusion} \label{sec:conclusion}
In this paper, we have proposed an \name framework to achieve the abnormal status monitoring in a vehicle for both driver and passengers. 
We have first proposed a novel adaptive threshold pseudo-labeling scheme to facilitate balanced class distribution for training with the enhanced feature extraction from other unlabeled data. Then, we have designed a missing modality reconstruction module to recover modality-specific features based on cross-modality relationships learned from limited labels.
Comparing with state-of-the-art baselines, our extensive evaluations have demonstrated that our \name achieves \needrev{outstanding performance and fast convergence}.

As a potential future direction, we are looking forward to
extending our \name to various cases such as
distributed learning systems~\cite{zhang2024satfed,zheng2023autofed,zhang2024fedac,lin2024adaptsfl,lin2024split,lyu2023optimal,lin2023fedsn,hu2024accelerating}, large language models~\cite{hu2024agentscodriver,lin2023pushing,fang2024automated,hu2024agentscomerge,lin2024splitlora,qiu2024ifvit} etc.


\ifCLASSOPTIONcaptionsoff
  \newpage
\fi



%



\bibliographystyle{IEEEtran}
\bibliography{reference}

\begin{thebibliography}{10}
\providecommand{\url}[1]{#1}
\csname url@samestyle\endcsname
\providecommand{\newblock}{\relax}
\providecommand{\bibinfo}[2]{#2}
\providecommand{\BIBentrySTDinterwordspacing}{\spaceskip=0pt\relax}
\providecommand{\BIBentryALTinterwordstretchfactor}{4}
\providecommand{\BIBentryALTinterwordspacing}{\spaceskip=\fontdimen2\font plus
\BIBentryALTinterwordstretchfactor\fontdimen3\font minus \fontdimen4\font\relax}
\providecommand{\BIBforeignlanguage}[2]{{%
\expandafter\ifx\csname l@#1\endcsname\relax
\typeout{** WARNING: IEEEtran.bst: No hyphenation pattern has been}%
\typeout{** loaded for the language `#1'. Using the pattern for}%
\typeout{** the default language instead.}%
\else
\language=\csname l@#1\endcsname
\fi
#2}}
\providecommand{\BIBdecl}{\relax}
\BIBdecl

\bibitem{Association}
Association for safe international road travel. accessed feb. 2015. Http://www.asirt.org/.

\bibitem{WHO}
Who: The top 10 causes of death. Https://www.who.int/news-room/fact-sheets/detail/the-top-10-causes-of-death.

\bibitem{dingus2016driver}
T.~A. Dingus, F.~Guo, S.~Lee, J.~F. Antin, M.~Perez, M.~Buchanan-King, and J.~Hankey, ``Driver crash risk factors and prevalence evaluation using naturalistic driving data,'' \emph{Proceedings of the National Academy of Sciences}, vol. 113, no.~10, pp. 2636--2641, 2016.

\bibitem{jegham2019mdad}
I.~Jegham, A.~Ben~Khalifa, I.~Alouani, and M.~A. Mahjoub, ``Mdad: A multimodal and multiview in-vehicle driver action dataset,'' in \emph{Computer Analysis of Images and Patterns: 18th International Conference, CAIP 2019, Salerno, Italy, September 3--5, 2019, Proceedings, Part I 18}.\hskip 1em plus 0.5em minus 0.4em\relax Springer, 2019, pp. 518--529.

\bibitem{rastgoo2018critical}
M.~N. Rastgoo, B.~Nakisa, A.~Rakotonirainy, V.~Chandran, and D.~Tjondronegoro, ``A critical review of proactive detection of driver stress levels based on multimodal measurements,'' \emph{ACM Computing Surveys (CSUR)}, vol.~51, no.~5, pp. 1--35, 2018.

\bibitem{zhang2016traffic}
G.~Zhang, K.~K. Yau, X.~Zhang, and Y.~Li, ``Traffic accidents involving fatigue driving and their extent of casualties,'' \emph{Accident Analysis \& Prevention}, vol.~87, pp. 34--42, 2016.

\bibitem{hamza2020monitoring}
F.~Hamza~Cherif, L.~Hamza~Cherif, M.~Benabdellah, and G.~Nassar, ``Monitoring driver health status in real time,'' \emph{Review of scientific instruments}, vol.~91, no.~3, 2020.

\bibitem{nvemcova2020multimodal}
A.~N{\v{e}}mcov{\'a}, V.~Svozilov{\'a}, K.~Bucsuh{\'a}zy, R.~Sm{\'\i}{\v{s}}ek, M.~M{\'e}zl, B.~Hesko, M.~Bel{\'a}k, M.~Bil{\'\i}k, P.~Maxera, M.~Seitl \emph{et~al.}, ``Multimodal features for detection of driver stress and fatigue,'' \emph{IEEE Transactions on Intelligent Transportation Systems}, vol.~22, no.~6, pp. 3214--3233, 2020.

\bibitem{kaplan2015driver}
S.~Kaplan, M.~A. Guvensan, A.~G. Yavuz, and Y.~Karalurt, ``Driver behavior analysis for safe driving: A survey,'' \emph{IEEE Transactions on Intelligent Transportation Systems}, vol.~16, no.~6, pp. 3017--3032, 2015.

\bibitem{juncen2023mmdrive}
Z.~Juncen, J.~Cao, Y.~Yang, W.~Ren, and H.~Han, ``mmdrive: Fine-grained fatigue driving detection using mmwave radar,'' \emph{ACM Transactions on Internet of Things}, vol.~4, no.~4, pp. 1--30, 2023.

\bibitem{pakdamanian2021deeptake}
E.~Pakdamanian, S.~Sheng, S.~Baee, S.~Heo, S.~Kraus, and L.~Feng, ``Deeptake: Prediction of driver takeover behavior using multimodal data,'' in \emph{Proceedings of the 2021 CHI Conference on Human Factors in Computing Systems}, 2021, pp. 1--14.

\bibitem{du2020vision}
G.~Du, T.~Li, C.~Li, P.~X. Liu, and D.~Li, ``Vision-based fatigue driving recognition method integrating heart rate and facial features,'' \emph{IEEE transactions on intelligent transportation systems}, vol.~22, no.~5, pp. 3089--3100, 2020.

\bibitem{alioua2014driver}
N.~Alioua, A.~Amine, and M.~Rziza, ``Driver’s fatigue detection based on yawning extraction,'' \emph{International journal of vehicular technology}, vol. 2014, 2014.

\bibitem{sigari2013driver}
M.-H. Sigari, M.~Fathy, and M.~Soryani, ``A driver face monitoring system for fatigue and distraction detection,'' \emph{International journal of vehicular technology}, vol. 2013, pp. 1--11, 2013.

\bibitem{yang2018car}
M.~Yang, X.~Yang, L.~Li, and L.~Zhang, ``In-car multiple targets vital sign monitoring using location-based vmd algorithm,'' in \emph{2018 10th International Conference on Wireless Communications and Signal Processing (WCSP)}.\hskip 1em plus 0.5em minus 0.4em\relax IEEE, 2018, pp. 1--6.

\bibitem{ghorbani2023self}
R.~Ghorbani, M.~J. Reinders, and D.~M. Tax, ``Self-supervised ppg representation learning shows high inter-subject variability,'' in \emph{Proceedings of the 2023 8th International Conference on Machine Learning Technologies}, 2023, pp. 127--132.

\bibitem{taamneh2017multimodal}
S.~Taamneh, P.~Tsiamyrtzis, M.~Dcosta, P.~Buddharaju, A.~Khatri, M.~Manser, T.~Ferris, R.~Wunderlich, and I.~Pavlidis, ``A multimodal dataset for various forms of distracted driving,'' \emph{Scientific data}, vol.~4, no.~1, pp. 1--21, 2017.

\bibitem{yang2023aide}
D.~Yang, S.~Huang, Z.~Xu, Z.~Li, S.~Wang, M.~Li, Y.~Wang, Y.~Liu, K.~Yang, Z.~Chen \emph{et~al.}, ``Aide: A vision-driven multi-view, multi-modal, multi-tasking dataset for assistive driving perception,'' in \emph{Proceedings of the IEEE/CVF International Conference on Computer Vision}, 2023, pp. 20\,459--20\,470.

\bibitem{hu2024collaborative}
S.~Hu, Z.~Fang, Y.~Deng, X.~Chen, and Y.~Fang, ``Collaborative perception for connected and autonomous driving: Challenges, possible solutions and opportunities,'' \emph{arXiv preprint arXiv:2401.01544}, 2024.

\bibitem{lin2024efficient}
Z.~Lin, G.~Zhu, Y.~Deng, X.~Chen, Y.~Gao, K.~Huang, and Y.~Fang, ``Efficient parallel split learning over resource-constrained wireless edge networks,'' \emph{IEEE Trans. Mob. Comput.}, 2024.

\bibitem{peng2024sums}
J.~Peng, Z.~Chen, Z.~Lin, H.~Yuan, Z.~Fang, L.~Bao, Z.~Song, Y.~Li, J.~Ren, and Y.~Gao, ``Sums: Sniffing unknown multiband signals under low sampling rates,'' \emph{arXiv preprint arXiv:2405.15705}, 2024.

\bibitem{yuan2024satsense}
H.~Yuan, Z.~Chen, Z.~Lin, J.~Peng, Z.~Fang, Y.~Zhong, Z.~Song, and Y.~Gao, ``Satsense: Multi-satellite collaborative framework for spectrum sensing,'' \emph{arXiv preprint arXiv:2405.15542}, 2024.

\bibitem{hu2023towards}
S.~Hu, Z.~Fang, X.~Chen, Y.~Fang, and S.~Kwong, ``Towards full-scene domain generalization in multi-agent collaborative bird's eye view segmentation for connected and autonomous driving,'' \emph{arXiv preprint arXiv:2311.16754}, 2023.

\bibitem{zhang2024satfed}
Y.~Zhang, Z.~Lin, Z.~Chen, Z.~Fang, W.~Zhu, X.~Chen, J.~Zhao, and Y.~Gao, ``Satfed: A resource-efficient leo satellite-assisted heterogeneous federated learning framework,'' \emph{arXiv preprint arXiv:2409.13503}, 2024.

\bibitem{lin2022channel}
Z.~Lin, L.~Wang, J.~Ding, B.~Tan, and S.~Jin, ``Channel power gain estimation for terahertz vehicle-to-infrastructure networks,'' \emph{IEEE Commun. Lett.}, vol.~27, no.~1, pp. 155--159, 2022.

\bibitem{hu2023adaptive}
S.~Hu, Z.~Fang, H.~An, G.~Xu, Y.~Zhou, X.~Chen, and Y.~Fang, ``Adaptive communications in collaborative perception with domain alignment for autonomous driving,'' \emph{arXiv preprint arXiv:2310.00013}, 2023.

\bibitem{bao2024bmad}
J.~Bao, H.~Sun, H.~Deng, Y.~He, Z.~Zhang, and X.~Li, ``Bmad: Benchmarks for medical anomaly detection,'' in \emph{Proceedings of the IEEE/CVF Conference on Computer Vision and Pattern Recognition}, 2024, pp. 4042--4053.

\bibitem{ghorbani2024personalized}
R.~Ghorbani, M.~J. Reinders, and D.~M. Tax, ``Personalized anomaly detection in ppg data using representation learning and biometric identification,'' \emph{Biomedical Signal Processing and Control}, vol.~94, p. 106216, 2024.

\bibitem{ma2021smil}
M.~Ma, J.~Ren, L.~Zhao, S.~Tulyakov, C.~Wu, and X.~Peng, ``Smil: Multimodal learning with severely missing modality,'' in \emph{Proceedings of the AAAI Conference on Artificial Intelligence}, vol.~35, no.~3, 2021, pp. 2302--2310.

\bibitem{wang2020multimodal}
Q.~Wang, L.~Zhan, P.~Thompson, and J.~Zhou, ``Multimodal learning with incomplete modalities by knowledge distillation,'' in \emph{Proceedings of the 26th ACM SIGKDD International Conference on Knowledge Discovery \& Data Mining}, 2020, pp. 1828--1838.

\bibitem{wang2016global}
H.~Wang, M.~Naghavi, C.~Allen, R.~M. Barber, Z.~A. Bhutta, A.~Carter, D.~C. Casey, F.~J. Charlson, A.~Z. Chen, M.~M. Coates \emph{et~al.}, ``Global, regional, and national life expectancy, all-cause mortality, and cause-specific mortality for 249 causes of death, 1980--2015: a systematic analysis for the global burden of disease study 2015,'' \emph{The lancet}, vol. 388, no. 10053, pp. 1459--1544, 2016.

\bibitem{fuster2018somatic}
J.~J. Fuster and K.~Walsh, ``Somatic mutations and clonal hematopoiesis: unexpected potential new drivers of age-related cardiovascular disease,'' \emph{Circulation research}, vol. 122, no.~3, pp. 523--532, 2018.

\bibitem{cho2011sector}
S.-H. Cho, Y.-Y. Nam, S.-J. Hong, and W.-D. Cho, ``Sector based scanning and adaptive active tracking of multiple objects,'' \emph{KSII Transactions on Internet and Information Systems (TIIS)}, vol.~5, no.~6, pp. 1166--1191, 2011.

\bibitem{lane1998tracking}
D.~M. Lane, M.~Chantler, D.~Y. Dai, and I.~T. Ruiz, ``Tracking and classification of multiple objects in multibeam sector scan sonar image sequences,'' in \emph{Proceedings of 1998 International Symposium on Underwater Technology}.\hskip 1em plus 0.5em minus 0.4em\relax IEEE, 1998, pp. 269--273.

\bibitem{perera2006multi}
A.~A. Perera, C.~Srinivas, A.~Hoogs, G.~Brooksby, and W.~Hu, ``Multi-object tracking through simultaneous long occlusions and split-merge conditions,'' in \emph{2006 IEEE Computer Society Conference on Computer Vision and Pattern Recognition (CVPR'06)}, vol.~1.\hskip 1em plus 0.5em minus 0.4em\relax IEEE, 2006, pp. 666--673.

\bibitem{yao2017deepsense}
S.~Yao, S.~Hu, Y.~Zhao, A.~Zhang, and T.~Abdelzaher, ``Deepsense: A unified deep learning framework for time-series mobile sensing data processing,'' in \emph{Proceedings of the 26th international conference on world wide web}, 2017, pp. 351--360.

\bibitem{suo2019metric}
Q.~Suo, W.~Zhong, F.~Ma, Y.~Yuan, J.~Gao, and A.~Zhang, ``Metric learning on healthcare data with incomplete modalities.'' in \emph{IJCAI}, vol. 3534, 2019, p. 3540.

\bibitem{jha2021multimodal}
S.~Jha, M.~F. Marzban, T.~Hu, M.~H. Mahmoud, N.~Al-Dhahir, and C.~Busso, ``The multimodal driver monitoring database: A naturalistic corpus to study driver attention,'' \emph{IEEE Transactions on Intelligent Transportation Systems}, vol.~23, no.~8, pp. 10\,736--10\,752, 2021.

\bibitem{ortega2020dmd}
J.~D. Ortega, N.~Kose, P.~Ca{\~n}as, M.-A. Chao, A.~Unnervik, M.~Nieto, O.~Otaegui, and L.~Salgado, ``Dmd: A large-scale multi-modal driver monitoring dataset for attention and alertness analysis,'' in \emph{Computer Vision--ECCV 2020 Workshops: Glasgow, UK, August 23--28, 2020, Proceedings, Part IV 16}.\hskip 1em plus 0.5em minus 0.4em\relax Springer, 2020, pp. 387--405.

\bibitem{angkititrakul2007utdrive}
P.~Angkititrakul, M.~Petracca, A.~Sathyanarayana, and J.~H. Hansen, ``Utdrive: Driver behavior and speech interactive systems for in-vehicle environments,'' in \emph{2007 IEEE Intelligent Vehicles Symposium}.\hskip 1em plus 0.5em minus 0.4em\relax IEEE, 2007, pp. 566--569.

\bibitem{lee2013pseudo}
D.-H. Lee \emph{et~al.}, ``Pseudo-label: The simple and efficient semi-supervised learning method for deep neural networks,'' in \emph{Workshop on challenges in representation learning, ICML}, vol.~3, no.~2.\hskip 1em plus 0.5em minus 0.4em\relax Atlanta, 2013, p. 896.

\bibitem{mclachlan1975iterative}
G.~J. McLachlan, ``Iterative reclassification procedure for constructing an asymptotically optimal rule of allocation in discriminant analysis,'' \emph{Journal of the American Statistical Association}, vol.~70, no. 350, pp. 365--369, 1975.

\bibitem{bachman2014learning}
P.~Bachman, O.~Alsharif, and D.~Precup, ``Learning with pseudo-ensembles,'' \emph{Advances in neural information processing systems}, vol.~27, 2014.

\bibitem{raffel2020exploring}
C.~Raffel, N.~Shazeer, A.~Roberts, K.~Lee, S.~Narang, M.~Matena, Y.~Zhou, W.~Li, and P.~J. Liu, ``Exploring the limits of transfer learning with a unified text-to-text transformer,'' \emph{Journal of machine learning research}, vol.~21, no. 140, pp. 1--67, 2020.

\bibitem{sohn2020fixmatch}
K.~Sohn, D.~Berthelot, N.~Carlini, Z.~Zhang, H.~Zhang, C.~A. Raffel, E.~D. Cubuk, A.~Kurakin, and C.-L. Li, ``Fixmatch: Simplifying semi-supervised learning with consistency and confidence,'' \emph{Advances in neural information processing systems}, vol.~33, pp. 596--608, 2020.

\bibitem{berthelot2019mixmatch}
D.~Berthelot, N.~Carlini, I.~Goodfellow, N.~Papernot, A.~Oliver, and C.~A. Raffel, ``Mixmatch: A holistic approach to semi-supervised learning,'' \emph{Advances in neural information processing systems}, vol.~32, 2019.

\bibitem{xie2020unsupervised}
Q.~Xie, Z.~Dai, E.~Hovy, T.~Luong, and Q.~Le, ``Unsupervised data augmentation for consistency training,'' \emph{Advances in neural information processing systems}, vol.~33, pp. 6256--6268, 2020.

\bibitem{zhang2021flexmatch}
B.~Zhang, Y.~Wang, W.~Hou, H.~Wu, J.~Wang, M.~Okumura, and T.~Shinozaki, ``Flexmatch: Boosting semi-supervised learning with curriculum pseudo labeling,'' \emph{Advances in Neural Information Processing Systems}, vol.~34, pp. 18\,408--18\,419, 2021.

\bibitem{guo2022class}
L.-Z. Guo and Y.-F. Li, ``Class-imbalanced semi-supervised learning with adaptive thresholding,'' in \emph{International conference on machine learning}.\hskip 1em plus 0.5em minus 0.4em\relax PMLR, 2022, pp. 8082--8094.

\bibitem{zhao2021missing}
J.~Zhao, R.~Li, and Q.~Jin, ``Missing modality imagination network for emotion recognition with uncertain missing modalities,'' in \emph{Proceedings of the 59th Annual Meeting of the Association for Computational Linguistics and the 11th International Joint Conference on Natural Language Processing (Volume 1: Long Papers)}, 2021, pp. 2608--2618.

\bibitem{zhou2017anomaly}
C.~Zhou and R.~C. Paffenroth, ``Anomaly detection with robust deep autoencoders,'' in \emph{Proceedings of the 23rd ACM SIGKDD international conference on knowledge discovery and data mining}, 2017, pp. 665--674.

\bibitem{zhou2021memorizing}
K.~Zhou, J.~Li, Y.~Xiao, J.~Yang, J.~Cheng, W.~Liu, W.~Luo, J.~Liu, and S.~Gao, ``Memorizing structure-texture correspondence for image anomaly detection,'' \emph{IEEE transactions on neural networks and learning systems}, vol.~33, no.~6, pp. 2335--2349, 2021.

\bibitem{perera2019ocgan}
P.~Perera, R.~Nallapati, and B.~Xiang, ``Ocgan: One-class novelty detection using gans with constrained latent representations,'' in \emph{Proceedings of the IEEE/CVF conference on computer vision and pattern recognition}, 2019, pp. 2898--2906.

\bibitem{liu2011latent}
G.~Liu and S.~Yan, ``Latent low-rank representation for subspace segmentation and feature extraction,'' in \emph{2011 international conference on computer vision}.\hskip 1em plus 0.5em minus 0.4em\relax IEEE, 2011, pp. 1615--1622.

\bibitem{rao2009motion}
S.~Rao, R.~Tron, R.~Vidal, and Y.~Ma, ``Motion segmentation in the presence of outlying, incomplete, or corrupted trajectories,'' \emph{IEEE transactions on pattern analysis and machine intelligence}, vol.~32, no.~10, pp. 1832--1845, 2009.

\bibitem{ma2007segmentation}
Y.~Ma, H.~Derksen, W.~Hong, and J.~Wright, ``Segmentation of multivariate mixed data via lossy data coding and compression,'' \emph{IEEE transactions on pattern analysis and machine intelligence}, vol.~29, no.~9, pp. 1546--1562, 2007.

\bibitem{labrin2020principal}
C.~Labr{\'\i}n and F.~Urdinez, ``{Principal Component Analysis},'' in \emph{R for political data science}, Nov. 2020, pp. 375--393.

\bibitem{goodfellow2014generative}
I.~Goodfellow, J.~Pouget-Abadie, M.~Mirza, B.~Xu, D.~Warde-Farley, S.~Ozair, A.~Courville, and Y.~Bengio, ``Generative adversarial nets,'' \emph{Advances in neural information processing systems}, vol.~27, 2014.

\bibitem{finn2017model}
C.~Finn, P.~Abbeel, and S.~Levine, ``Model-agnostic meta-learning for fast adaptation of deep networks,'' in \emph{International conference on machine learning}.\hskip 1em plus 0.5em minus 0.4em\relax PMLR, 2017, pp. 1126--1135.

\bibitem{gui2018few}
L.-Y. Gui, Y.-X. Wang, D.~Ramanan, and J.~M. Moura, ``Few-shot human motion prediction via meta-learning,'' in \emph{Proceedings of the European Conference on Computer Vision (ECCV)}, 2018, pp. 432--450.

\bibitem{sun2019meta}
Q.~Sun, Y.~Liu, T.-S. Chua, and B.~Schiele, ``Meta-transfer learning for few-shot learning,'' in \emph{Proceedings of the IEEE/CVF conference on computer vision and pattern recognition}, 2019, pp. 403--412.

\bibitem{chen2021meta}
Y.~Chen, Z.~Liu, H.~Xu, T.~Darrell, and X.~Wang, ``Meta-baseline: Exploring simple meta-learning for few-shot learning,'' in \emph{Proceedings of the IEEE/CVF international conference on computer vision}, 2021, pp. 9062--9071.

\bibitem{martin2019drive}
M.~Martin, A.~Roitberg, M.~Haurilet, M.~Horne, S.~Rei{\ss}, M.~Voit, and R.~Stiefelhagen, ``Drive\&act: A multi-modal dataset for fine-grained driver behavior recognition in autonomous vehicles,'' in \emph{Proceedings of the IEEE/CVF International Conference on Computer Vision}, 2019, pp. 2801--2810.

\bibitem{tian2020contrastive}
Y.~Tian, D.~Krishnan, and P.~Isola, ``Contrastive multiview coding,'' in \emph{Computer Vision--ECCV 2020: 16th European Conference, Glasgow, UK, August 23--28, 2020, Proceedings, Part XI 16}, 2020, pp. 776--794.

\bibitem{li2021novel}
Z.~Li, L.~Chen, L.~Nie, and S.~X. Yang, ``A novel learning model of driver fatigue features representation for steering wheel angle,'' \emph{IEEE Transactions on Vehicular Technology}, vol.~71, no.~1, pp. 269--281, 2021.

\bibitem{abedi2022use}
H.~Abedi, M.~Ma, J.~Yu, J.~He, A.~Ansariyan, and G.~Shaker, ``On the use of machine learning and deep learning for radar-based passenger monitoring,'' in \emph{2022 IEEE International Symposium on Antennas and Propagation and USNC-URSI Radio Science Meeting (AP-S/URSI)}.\hskip 1em plus 0.5em minus 0.4em\relax IEEE, 2022, pp. 902--903.

\bibitem{breed2009dynamic}
D.~S. Breed, W.~E. DuVall, and W.~C. Johnson, ``Dynamic weight sensing and classification of vehicular occupants,'' Nov.~17 2009, uS Patent 7,620,521.

\bibitem{backar2022vehicle}
L.~H. Backar, M.~A. Khalifa, and M.~A.-M. Salem, ``In-vehicle monitoring for passengers' safety,'' in \emph{2022 IEEE 12th International Conference on Consumer Electronics (ICCE-Berlin)}, 2022, pp. 1--6.

\bibitem{ma2020carosense}
Y.~Ma, Y.~Zeng, and V.~Jain, ``Carosense: Car occupancy sensing with the ultra-wideband keyless infrastructure,'' \emph{Proceedings of the ACM on Interactive, Mobile, Wearable and Ubiquitous Technologies}, vol.~4, no.~3, pp. 1--28, 2020.

\bibitem{wang2022freematch}
Y.~Wang, H.~Chen, Q.~Heng, W.~Hou, Y.~Fan, Z.~Wu, J.~Wang, M.~Savvides, T.~Shinozaki, B.~Raj \emph{et~al.}, ``Freematch: Self-adaptive thresholding for semi-supervised learning,'' \emph{arXiv preprint arXiv:2205.07246}, 2022.

\bibitem{wang2020transmodality}
Z.~Wang, Z.~Wan, and X.~Wan, ``Transmodality: An end2end fusion method with transformer for multimodal sentiment analysis,'' in \emph{Proceedings of the web conference 2020}, 2020, pp. 2514--2520.

\bibitem{zheng2023autofed}
T.~Zheng, A.~Li, Z.~Chen, H.~Wang, and J.~Luo, ``Autofed: Heterogeneity-aware federated multimodal learning for robust autonomous driving,'' in \emph{Proceedings of the 29th Annual International Conference on Mobile Computing and Networking}, 2023, pp. 1--15.

\bibitem{zhang2024fedac}
Y.~Zhang, H.~Chen, Z.~Lin, Z.~Chen, and J.~Zhao, ``Fedac: A adaptive clustered federated learning framework for heterogeneous data,'' \emph{arXiv preprint arXiv:2403.16460}, 2024.

\bibitem{lin2024adaptsfl}
Z.~Lin, G.~Qu, W.~Wei, X.~Chen, and K.~K. Leung, ``Adaptsfl: Adaptive split federated learning in resource-constrained edge networks,'' \emph{arXiv preprint arXiv:2403.13101}, 2024.

\bibitem{lin2024split}
Z.~Lin, G.~Qu, X.~Chen, and K.~Huang, ``Split learning in 6g edge networks,'' \emph{IEEE Wirel. Commun.}, 2024.

\bibitem{lyu2023optimal}
S.~Lyu, Z.~Lin, G.~Qu, X.~Chen, X.~Huang, and P.~Li, ``Optimal resource allocation for u-shaped parallel split learning,'' in \emph{Proc. Globecom Wkshps}, 2023, pp. 197--202.

\bibitem{lin2023fedsn}
Z.~Lin, Z.~Chen, Z.~Fang, X.~Chen, X.~Wang, and Y.~Gao, ``Fedsn: A general federated learning framework over leo satellite networks,'' \emph{arXiv preprint arXiv:2311.01483}, 2023.

\bibitem{hu2024accelerating}
M.~Hu, J.~Zhang, X.~Wang, S.~Liu, and Z.~Lin, ``Accelerating federated learning with model segmentation for edge networks,'' \emph{IEEE Transactions on Green Communications and Networking}, 2024.

\bibitem{hu2024agentscodriver}
S.~Hu, Z.~Fang, Z.~Fang, X.~Chen, and Y.~Fang, ``Agentscodriver: Large language model empowered collaborative driving with lifelong learning,'' \emph{arXiv preprint arXiv:2404.06345}, 2024.

\bibitem{lin2023pushing}
Z.~Lin, G.~Qu, Q.~Chen, X.~Chen, Z.~Chen, and K.~Huang, ``Pushing large language models to the 6g edge: Vision, challenges, and opportunities,'' \emph{arXiv preprint arXiv:2309.16739}, 2023.

\bibitem{fang2024automated}
Z.~Fang, Z.~Lin, Z.~Chen, X.~Chen, Y.~Gao, and Y.~Fang, ``Automated federated pipeline for parameter-efficient fine-tuning of large language models,'' \emph{arXiv preprint arXiv:2404.06448}, 2024.

\bibitem{hu2024agentscomerge}
S.~Hu, Z.~Fang, Z.~Fang, Y.~Deng, X.~Chen, Y.~Fang, and S.~Kwong, ``Agentscomerge: Large language model empowered collaborative decision making for ramp merging,'' \emph{arXiv preprint arXiv:2408.03624}, 2024.

\bibitem{lin2024splitlora}
Z.~Lin, X.~Hu, Y.~Zhang, Z.~Chen, Z.~Fang, X.~Chen, A.~Li, P.~Vepakomma, and Y.~Gao, ``Splitlora: A split parameter-efficient fine-tuning framework for large language models,'' \emph{arXiv preprint arXiv:2407.00952}, 2024.

\bibitem{qiu2024ifvit}
Y.~Qiu, H.~Chen, X.~Dong, Z.~Lin, I.~Y. Liao, M.~Tistarelli, and Z.~Jin, ``Ifvit: Interpretable fixed-length representation for fingerprint matching via vision transformer,'' \emph{arXiv preprint arXiv:2404.08237}, 2024.

\end{thebibliography}

\end{document}